\documentclass[10pt,twocolumn,letterpaper]{article}

\usepackage[pagenumbers]{cvpr} % To force page numbers, e.g. for an arXiv version

\usepackage{fix-cm}
\usepackage{xspace}
\usepackage[accsupp]{axessibility}
\usepackage[misc]{ifsym} %for \Letter

\usepackage{xcolor}
\usepackage{pifont}
\usepackage{colortbl}
\usepackage[ruled,vlined]{algorithm2e}

\newcommand{\methodname}{{GlyphPrinter}\xspace}
\newcommand{\datasetname}{{GlyphCorrector}\xspace}

\newcommand{\dpomethodname}{{R-GDPO}\xspace}

\newcommand{\completeinfermethodname}{{Regional Reward Guidance}\xspace}
\newcommand{\infermethodname}{{RRG}\xspace}

\newcommand{\multilingualbenchmarkname}{{GlyphAcc-Multilingual}\xspace}
\newcommand{\complexbenchmarkname}{{GlyphAcc-Complex}\xspace}

\definecolor{myorange}{RGB}{255,100,3}
\definecolor{mygray}{gray}{.85}
\definecolor{mygray1}{gray}{.7}
\definecolor{mygray2}{gray}{.93}
\definecolor{mygray3}{gray}{.90}

\newcommand{\myparagraph}[1]{{\vspace{.3em} \noindent \bf #1}}
\definecolor{cvprblue}{rgb}{0.21,0.49,0.74}
\usepackage[pagebackref,breaklinks,colorlinks,allcolors=cvprblue]{hyperref}

\def\paperID{2757} % *** Enter the Paper ID here
\def\confName{CVPR}
\def\confYear{2026}

\title{\methodname: Region-Grouped Direct Preference Optimization for Glyph-Accurate Visual Text Rendering}

\author{
Xincheng Shuai$^1$\footnotemark[1]
\quad
Ziye Li$^1$\footnotemark[1]
\quad
Henghui Ding$^1$~\!$^{\textrm{\Letter}}$
\quad
Dacheng Tao$^2$
\\
{\fontsize{11}{11}\selectfont $^1$Institute of Big Data, College of Computer Science and Artificial Intelligence, Fudan University, China}\\
{\fontsize{11}{11}\selectfont $^2$Generative AI Lab, College of Computing and Data Science, Nanyang Technological University, Singapore}
\\
{\tt\footnotesize henghui.ding@gmail.com\quad dacheng.tao@gmail.com}
\\
\href{https://henghuiding.com/GlyphPrinter/}{https://henghuiding.com/GlyphPrinter/}
}

\begin{document}

\twocolumn[{% 
\renewcommand\twocolumn[1][]{#1}% 
% \maketitle 
\maketitle
\vspace{-7.6mm}
\begin{center} 
\centering 
\captionsetup{type=figure}
\includegraphics[width=1\linewidth]{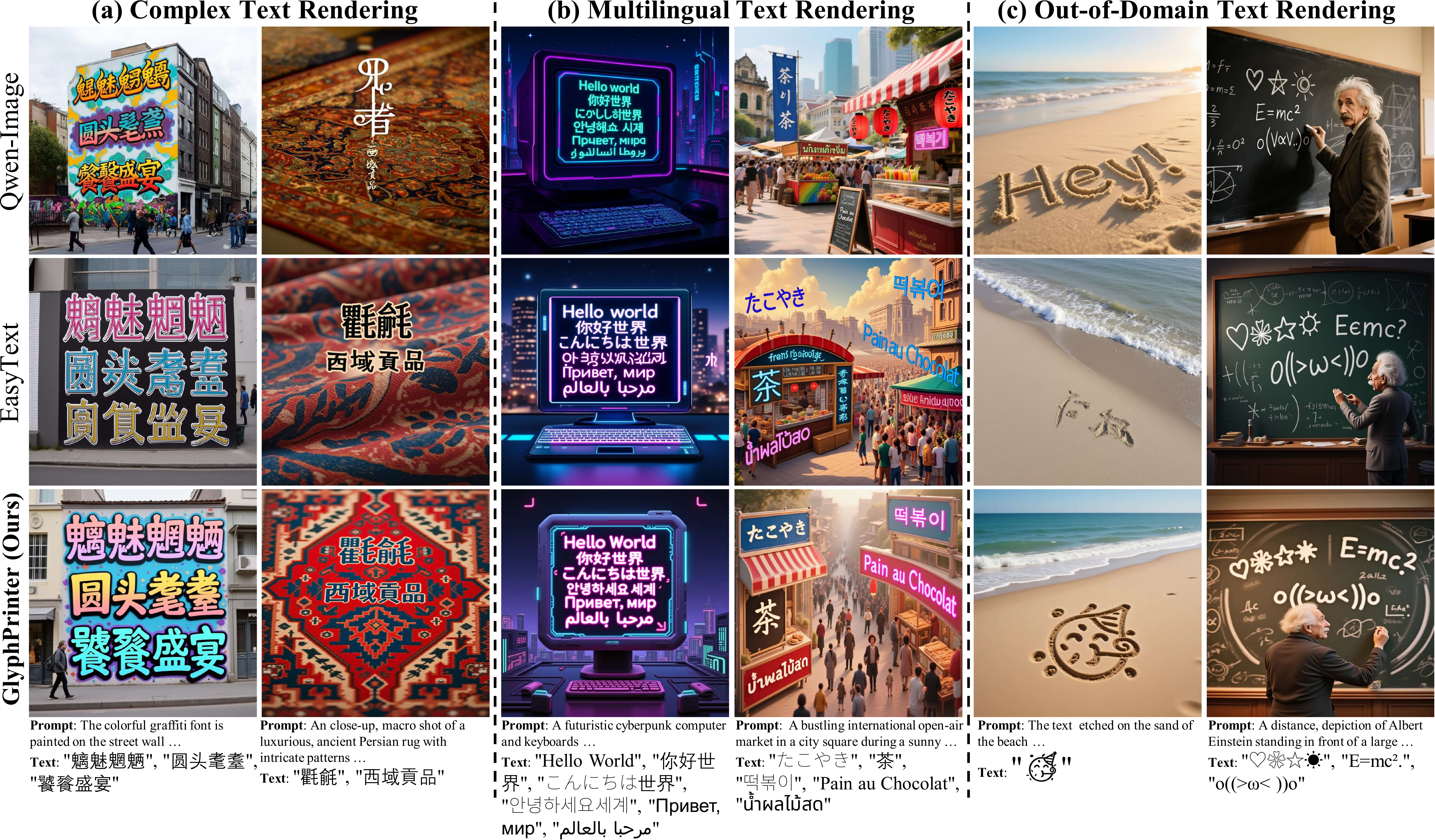}
\vspace{-6.6mm}
\captionof{figure}{Comparison of different methods on text rendering. The figures show that the proposed \methodname achieves higher glyph accuracy than advanced methods~\cite{wu2025qwen,lu2025easytext} in (\textbf{a}) complex, (\textbf{b}) multilingual, and (\textbf{c}) out-of-domain text rendering.}
\vspace{3.6mm}
\label{fig:teaser}
\end{center}
}]

\maketitle
\renewcommand{\thefootnote}{\fnsymbol{footnote}}
\footnotetext[1]{Equal contribution.}
\footnotetext[0]{${\textrm{\Letter}}$ Corresponding author (henghui.ding@gmail.com).}

\begin{abstract}
Generating accurate glyphs for visual text rendering is essential yet challenging. Existing methods typically enhance text rendering by training on a large amount of high-quality scene text images, but the limited coverage of glyph variations and excessive stylization often compromise glyph accuracy, especially for complex or out-of-domain characters. Some methods leverage reinforcement learning to alleviate this issue, yet their reward models usually depend on text recognition systems that are insensitive to fine-grained glyph errors, so images with incorrect glyphs may still receive high rewards. Inspired by Direct Preference Optimization (DPO), we propose \textbf{\methodname}, a preference-based text rendering method that eliminates reliance on explicit reward models. However, the standard DPO objective only models overall preference between two samples, which is insufficient for visual text rendering where glyph errors typically occur in localized regions. To address this issue, we construct the \textbf{\datasetname} dataset with region-level glyph preference annotations and propose {Region-Grouped DPO} (\textbf{R-GDPO}), a region-based objective that optimizes inter- and intra-sample preferences over annotated regions, substantially enhancing glyph accuracy. Furthermore, we introduce \textbf{Regional Reward Guidance}, an inference strategy that samples from an optimal distribution with controllable glyph accuracy. Extensive experiments demonstrate that the proposed \methodname outperforms existing methods in glyph accuracy while maintaining a favorable balance between stylization and precision.
\end{abstract}

\section{Introduction}
Visual text rendering has emerged as a critical task in image generation~\cite{rombach2022high,shuai2024survey,podell2023sdxl,shuai2025free2}. However, generating accurate glyphs still remains a challenge. Recent studies~\cite{yang2023glyphcontrol,wang2025uniglyph,wang2025designdiffusion,li2024joytype,peng2025bizgen} have improved the text rendering capabilities of text-to-image (T2I) models~\cite{rombach2022high,esser2024scaling}. However, as shown in \cref{fig:teaser}, some advanced methods still struggle to generate accurate glyphs in challenging scenarios, \eg, complex Chinese characters or emojis, limiting their applicability.

\begin{figure}[t]
    \centering
    \includegraphics[width=0.47\textwidth]{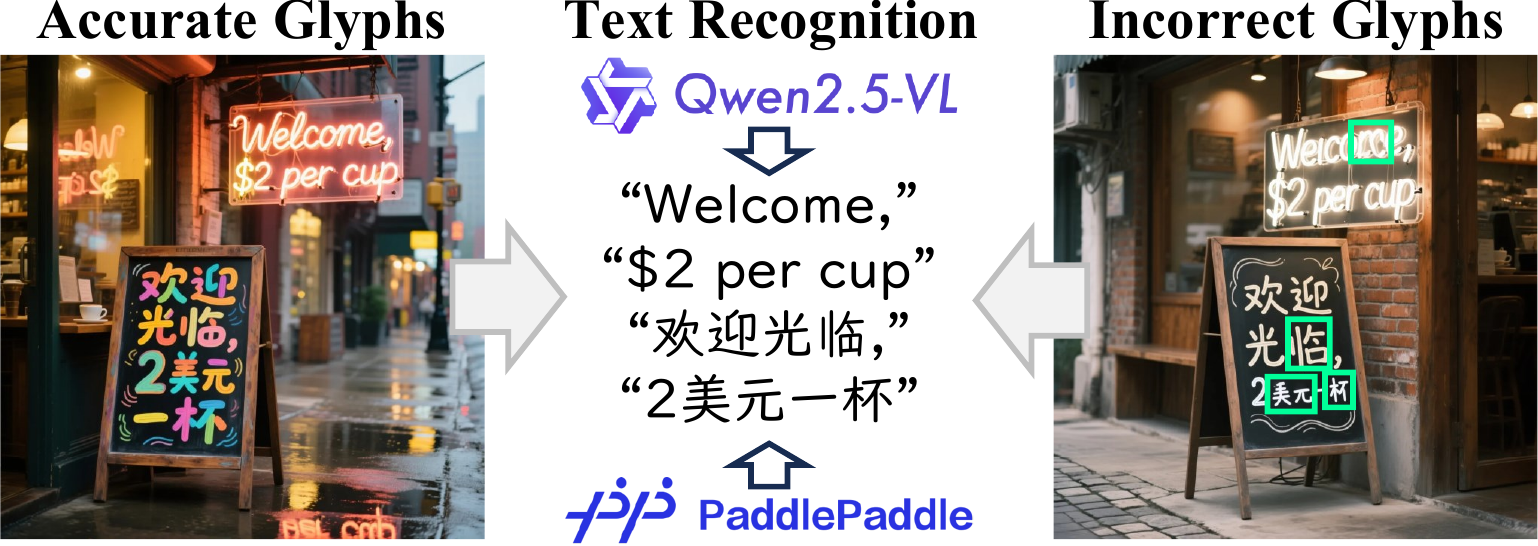}
    \vspace{-3.8mm}
    \caption{Text recognition models~\cite{du2020pp,wang2024qwen2} used in existing RL-based methods are insensitive to incorrect glyphs (highlighted with green boxes), leading to inflated rewards for such samples.}
    \label{fig:ocr_challenge}
    \vspace{-3mm}
\end{figure}

Based on how the glyph condition is injected, existing text rendering methods can be grouped into two categories. \textbf{1)} {Prompt-guided} methods~\cite{du2025textcrafter, wu2025qwen,geng2025x}, which are solely conditioned on embeddings encoded by character-level~\cite{liu2024glyph,liu2024glyphv2} or multilingual text encoders~\cite{gao2025seedream,wu2025qwen}.~\textbf{2)} {Glyph-image-guided} methods~\cite{jiang2025controltext, lan2025flux,lu2025easytext,tuo2023anytext}, which leverage rendered glyph images as a condition to better generate out-of-vocabulary characters. To achieve a better trade-off between stylization and glyph accuracy, some methods~\cite{wu2025qwen,gao2025seedream,geng2025x} train text rendering models on large-scale, high-quality scene text images. However, the limited coverage of glyph variations and excessive stylization compromise glyph accuracy, making it difficult to render complex or infrequent/out-of-domain glyphs. As shown in \cref{fig:teaser}, the generated images struggle to preserve structural details of the target characters, leading to extraneous or missing strokes. To alleviate this issue, some methods~\cite{geng2025x,liu2025flow} employ reinforcement learning (RL)~\cite{guo2025deepseek,rafailov2023direct} to improve model performance. They typically use OCR (Optical Character Recognition) models~\cite{du2020pp} or MLLMs (Multimodal Large Language Models)~\cite{wang2024qwen2,chen2024internvl} to assess the accuracy of generated texts and maximize a designed reward function. However, most of these text recognition models are insensitive to glyph errors. As shown in \cref{fig:ocr_challenge}, where both correct and incorrect text images are provided, they output identical predictions, leading to inflated rewards for samples with incorrect glyphs, ultimately impairing model performance.

These limitations motivate us to optimize the model using human preference data rather than recognition-based reward models. Inspired by Direct Preference Optimization (DPO)~\cite{wallace2024diffusion}, we propose \textbf{{\textit{\methodname}}}, a preference-based text rendering framework that no longer relies on explicit reward models. Standard DPO aligns model outputs with preferred images while discouraging inferior ones, and has mainly been used to improve image-level metrics, such as image quality. However, this image-level objective is suboptimal for text rendering, where glyph errors are typically localized. For example, given the glyph condition ``12345'', an image with the correct ``123'' and another with the correct ``45'' are selected as a pair. These images are then labeled as the winning and losing samples, respectively, since the former has higher overall accuracy ($60\% > 40\%$). Despite these image-level preferences, the model still struggles to learn accurate glyphs for ``45''.

To address this issue, we construct a region-level preference dataset, \textbf{\textit{\datasetname}}, where both correct and incorrect glyph regions are annotated for each image to define preference pairs. Specifically, we build a group of candidate images for each glyph condition, resulting in 7,117 images across 897 glyph conditions. Based on this dataset, we propose {Region-Grouped DPO} (\textbf{{R-GDPO}}), a region-based objective that leverages both inter- and intra-sample preference pairs: the former builds winning–losing pairs for the same region across different images within a group, while the latter uses correct and incorrect regions from the same image as preference pairs. R-GDPO enables more efficient data utilization and significantly improves glyph accuracy. In addition, we introduce Regional Reward Guidance (\textbf{\infermethodname}) to further enhance performance by sampling from an optimal distribution with controllable glyph accuracy. In summary, our main contributions are:
\begin{itemize}
\item We propose {{\methodname}}, a preference-based text rendering framework that improves glyph accuracy.

\item We construct {{\datasetname}}, a preference dataset with region-level annotations, facilitating the model to learn localized glyph correctness.

\item We design {R-GDPO} to fully exploit \datasetname by aligning model outputs with accurate glyph regions while distancing from incorrect ones, and introduce RRG to provide more controllability during inference.

\item Extensive experiments demonstrate that the proposed \methodname surpasses existing text rendering methods in glyph accuracy, while maintaining high image quality.

\end{itemize}

\section{Related Works}
\subsection{Visual Text Rendering}
Visual text rendering has become an increasingly important task in image generation~\cite{rombach2022high,podell2023sdxl,qin2025scenedesigner,li2025anyi2v,shuai2025free}. Existing approaches can be categorized into {prompt-guided}~\cite{liu2024glyph,chen2023textdiffuser,chen2024textdiffuser} and {glyph-image-guided}~\cite{wang2025reptext,xie2025textflux,gao2025postermaker} methods. The former ones, such as Glyph-Byt5-v2~\cite{liu2024glyphv2} and Seedream~\cite{gao2025seedream}, leverage character-level or multilingual text encoders to encode glyph information. These approaches require a large amount of text images for learning the mapping from the textual modality to visual glyphs. Additionally, they struggle to generate out-of-vocabulary characters, resulting in limited generalization capability. In contrast, {glyph-image-guided} methods~\cite{ma2025glyphdraw2,lan2025flux,tuo2023anytext}, including GlyphDraw~\cite{ma2023glyphdraw} and AnyText2~\cite{tuo2024anytext2}, alleviate this challenge by using glyph images rendered with a default font, but are susceptible to rigid/copy-and-paste style. Besides, most methods fail to produce accurate glyphs.

\subsection{Preference Alignment for Image Generation}
In large language models, Reinforcement Learning from Human Feedback (RLHF) improves the alignment of the model outputs with human preferences~\cite{rafailov2023direct,guo2025deepseek}. Recently, some studies have introduced the techniques into image generation to enhance the model's performance in downstream tasks~\cite{prabhudesai2023aligning,xu2023imagereward,black2023training,fan2023dpok,huang2025patchdpo,wallace2024diffusion}, like visual text rendering~\cite{geng2025x,liu2025flow}. However, the reward functions used in existing text rendering methods often rely on text recognition models~\cite{du2020pp,wang2024qwen2} that are insensitive to glyph errors. This can lead to inflated reward values being assigned to inferior images, impairing the performance.

\section{Preliminaries}
\myparagraph{Classifier-Free Guidance.} Instead of relying on an external classifier~\cite{dhariwal2021diffusion} to guide the sampling process, Classifier-Free Guidance (CFG)~\cite{ho2022classifier} trains the model with both conditional and unconditional inputs,  playing a crucial role in image generation. From the perspective of the score function, the sampling process of CFG can be expressed as:
\begin{equation}
\begin{aligned}\label{eq:cfg}
\hat{s}(x_t,\omega,\mathcal{C})&=(1-\omega)s(x_t)+\omega s(x_t,\mathcal{C}) \\
&=(1-\omega)\nabla_{x_t}\text{log}p(x_t)+\omega\nabla_{x_t}\text{log}p(x_t|\mathcal{C}) \\
&=\nabla_{x_t}\text{log}\big(p(x_t)^{(1-\omega)}p(x_t|\mathcal{C})^\omega\big), \\
\end{aligned}
\end{equation}
where $\mathcal{C}$ is task-specific condition, $s(x_t)$ and $s(x_t,\mathcal{C})$ are the unconditional and conditional score functions, respectively. The guidance weight $\omega$ can be adjusted to control the alignment of the generated images with input conditions.

\myparagraph{Reinforcement Learning from Human Feedback.}
Formally, Reinforcement Learning (RL) methods have the following objective:
\begin{multline}\label{eq:rl_objective}
\underset{p_\theta}{\text{min}}-E_{c\sim p_c,x_0 \sim  p_{\theta}(x_0|c)}[r(x_0,c)] \\
+\beta\mathcal{D}_{\text{KL}}\big(p_{\theta}(x_0|c)||p_\text{ref}(x_0|c)\big),
\end{multline}
where the reward model $r$ evaluates the generated image $x_0$ under the condition $c$. $p_\text{ref}$ and $p_\theta$ are the reference model and the one to be optimized. $\beta$ is the weight of KL-divergence $\mathcal{D}_{\text{KL}}$ for regularization. Then, \cref{eq:rl_objective} derives the global optimal solution $p_\theta^*(x_0|c)=p_\text{ref}(x_0|c)\text{exp}(r(x_0,c)/\beta)/Z(c)$, where the partition function $Z(c)$ takes the form $\Sigma_{x_0}p_\text{ref}(x_0|c)\text{exp}(r(x_0,c)/\beta)$.

For diffusion models in image generation, the RL methods~\cite{black2023training,guo2025deepseek,liu2025flow} using policy gradient are expensive and prone to reward hacking. In contrast, methods~\cite{huang2025patchdpo,wallace2024diffusion} optimized with DPO loss outperform in efficient and stable learning, which learn an implicit reward model~\cite{wallace2024diffusion}:
\begin{equation}\label{eq:dpo_reward}
% \resizebox{0.43\textwidth}{!}{$
r(x_0,c)=\beta T E_{t,x_{t-1,t}}\Big[\text{log}\frac{p_\theta^*(x_{t-1}|x_{t},c)}{p_\text{ref}(x_{t-1}|x_{t},c)}\Big]+\beta \text{log}Z(c),
% $}
\end{equation}
where $p_\theta^*$ is the optimal solution of \cref{eq:rl_objective}, and $x_{t-1,t}$ is sampled from $p_\theta^*(x_{t-1,t}|x_0,c)$. By maximizing the likelihood objective of the Bradly-Terry Model, the DPO loss for flow matching models is expressed as~\cite{wallace2024diffusion,liu2025improving}:
\begin{equation}
\begin{aligned}\label{eq:dpo_loss}
L_\text{DPO}&=-E_{c,x_0^w,x_0^l}[\text{log}\sigma(r(x_0^w,c)-r(x_0^l,c))] \\
&\leq -E_{c,t,x_0^w,x_0^l,\epsilon^w,\epsilon^l}\Big[\text{log}\sigma\big( -\beta T\omega_t(\\
&||v^w-v_\theta(x_t^w,t,c)||_2^2- ||v^w  -v_\text{ref}(x_t^w,t,c)||_2^2 \\
&-(||v^l-v_\theta(x_t^l,t,c)||_2^2 -||v^l-v_\text{ref}(x_t^l,t,c)||_2^2)) \big) \Big],
\end{aligned}
\end{equation}
where $\epsilon^w,\epsilon^l\sim \mathcal{N}(\mathbf{0},\mathbf{I})$, $x_0^w$ and $x_0^l$ are winning and losing samples from the preference dataset. The velocity $v$ takes the form $\epsilon-x_0$, and $x_t=(1-t)x_0+t\epsilon$. $\omega_t$ is the weight for each timestep. $v_\theta$ and the reference model $v_\text{ref}$ predict the velocity fields. It's worth noting that the partition function $Z(c)$ from \cref{eq:dpo_reward} cancels for preference pairs.

\section{Method: \methodname}
\begin{figure*}
    \centering
    \includegraphics[width=1\linewidth]{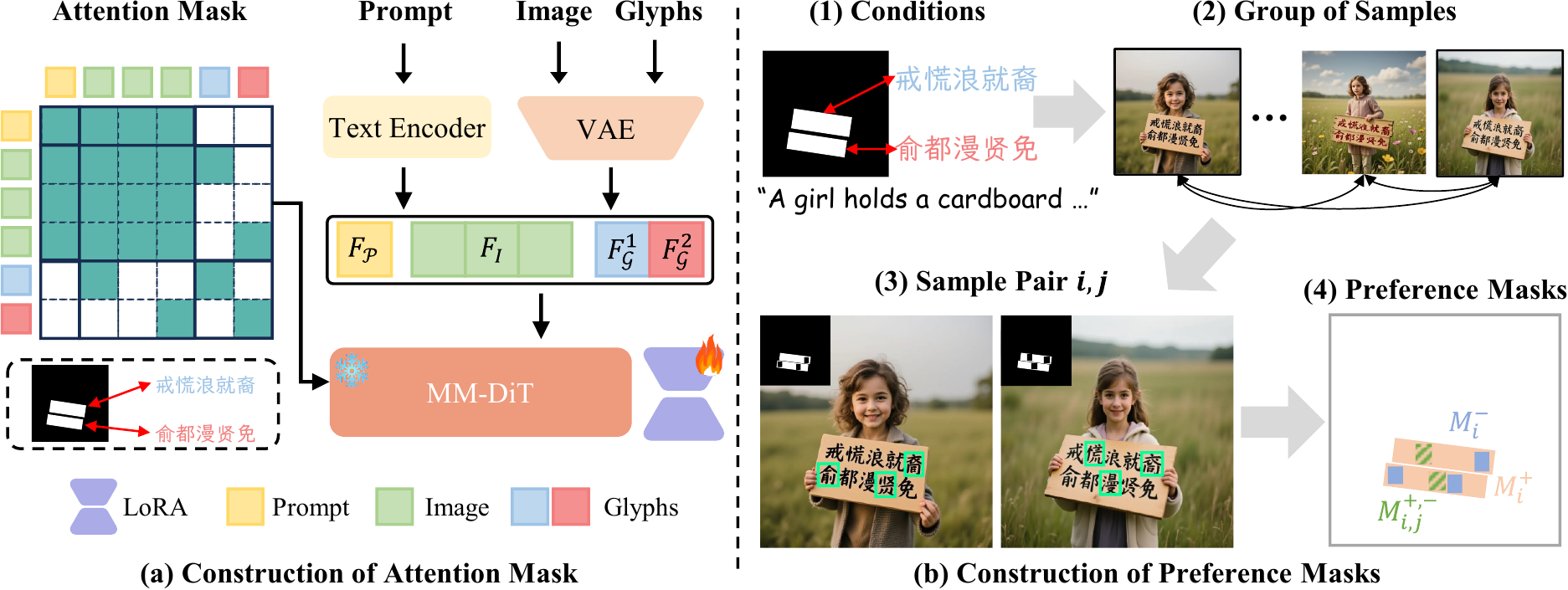}
    \vspace{-6mm}
    \caption{\textbf{(a) Construction of the attention mask used in \methodname}. In addition to the prompt-image and intra-modality attentions, we only enable the communication between the image features from the text region and the corresponding glyph feature for each text block. \textbf{(b) Construction of the preference masks used in \dpomethodname}. Our \datasetname contains region-level preference annotations for samples from each generated group, where the incorrect text regions are highlighted with green boxes. To make more efficient use of data, we simultaneously use \textit{inter-sample} ($M_{i,j}^{+,-}$) and \textit{intra-sample preference masks} ($M_{i}^{+},M_{i}^{-}$) to construct winning-losing pairs.}
    \label{fig:pipeline}

\end{figure*}

Given the prompt $\mathcal{P}$ and a set of $N_t$ text blocks $\left\{(\mathcal{T}_i,\mathbf{P}_i)\right\}_{i=1}^{N_t}$, where $\mathcal{T}$ and $\mathbf{P}$ denote the text to be rendered and its corresponding position, our text rendering model aims to generate the image that aligns with conditions. Specifically, \methodname first fine-tunes the T2I model~\cite{black2024flux} on text images to improve the text rendering ability, obtaining the baseline model. Then, \methodname is optimized with the proposed Region-Grouped Direct Preference Optimization (R-GDPO) on the preference dataset \datasetname for enhancing glyph accuracy. In addition, we introduce Regional Reward Guidance (RRG) during inference for more controllability. 

\subsection{Stage 1: Fine-Tuning on Text Images}
To improve the text rendering capability of the pre-trained T2I model~\cite{black2024flux}, we first fine-tune the model on text images, obtaining our baseline model. Specifically, $\mathcal{T}_i$ from each text block is first rendered to the glyph image $\mathcal{G}_i$ and then encoded by VAE~\cite{black2024flux}, which are assembled to compose the total glyph feature $\mathbf{F}_{\mathcal{G}}=\left\{\mathbf{F}_{\mathcal{G}}^i\right\}_{i=1}^{N_t}$. Then, we concatenate the prompt embedding $\mathbf{F}_{\mathcal{P}}$, noisy image $\mathbf{F}_{{I}}$, and $\mathbf{F}_{\mathcal{G}}$ along the sequence dimension, resulting in a joint representation $\mathbf{F} \in \mathbb{R}^{N\times D}$, where $N = N_\mathcal{P} + N_{I} + N_\mathcal{G}$ is the total sequence length and $D$ is the embedding dimension.

Then, we construct the attention mask $\mathcal{A}$ for text localization, as shown in \cref{fig:pipeline} (a). First, we denote the token ranges of $\mathbf{F}_{\mathcal{P}}$, $\mathbf{F}_{{I}}$, and $\mathbf{F}_{\mathcal{G}}$ as $\mathbf{R}_{\mathcal{P}}$, $\mathbf{R}_{I}$, and $\mathbf{R}_{\mathcal{G}}$, respectively. Furthermore, $\mathbf{R}_{\mathcal{G}}^i$ represents the range of glyph features from the $i$-th text block. Next, we can calculate the $(q,k)$-index element from $\mathcal{A}$ as:
\begin{equation}
\mathcal{A}[q,k] =
\begin{cases}\label{eq:attention_mask}
1, & q \in \mathbf{R}_{\mathcal{P}}, \; k \in \mathbf{R}_{I} \;\; \text{or} \;\; q \in \mathbf{R}_{I}, \; k \in \mathbf{R}_{\mathcal{P}} , \\
1, & q,k \in \mathbf{R}_{\mathcal{P}} \;\; \text{or} \;\; q,k \in \mathbf{R}_{I} \;\; \text{or} \;\; q,k \in \mathbf{R}_{\mathcal{G}}^i, \\
1, & q \in \mathbf{P}_i, \; k \in \mathbf{R}_{\mathcal{G}}^i \;\; \text{or} \;\; q \in \mathbf{R}_{\mathcal{G}}^i, \; k \in  \mathbf{P}_i, \\
0, & \text{otherwise.}
\end{cases}
\end{equation}
As indicated in \cref{eq:attention_mask} and \cref{fig:pipeline} (a), besides the prompt-image (1st row of \cref{eq:attention_mask}) and intra-modality attentions (2nd row), we only enable the communication between the region $\mathbf{P}_i$ in the image feature $\mathbf{F}_{{I}}$ and its corresponding glyph feature $\mathbf{F}_{\mathcal{G}}^i$ for each text block (3rd row).

Similar to previous studies~\cite{lu2025easytext,wu2025qwen,liu2024glyph,liu2024glyphv2}, our \methodname is first trained on multilingual synthetic text images and then fine-tuned on high-quality realistic data. The training objective of this stage is formulated as flow matching loss:
\begin{equation}
\begin{aligned}\label{eq:flow_loss}
E_{t,(x_0,c)}\left[\left\|v-v_{\theta}\left(x_t, t ,c\right)\right\|^2 \right],
\end{aligned}
\end{equation}
where $x_0$ is the text image, and the condition $c$ contains the prompt $\mathcal{P}$ and text blocks $\left\{(\mathcal{T}_i,\mathbf{P}_i)\right\}_{i=1}^{N_t}$.

\subsection{Stage 2: Region-Level Preference Optimization}
The model from Stage~1 can render stylized text at designated locations, but its glyph accuracy remains suboptimal, especially for complex or out-of-domain characters, mainly due to the limited glyph coverage of the training data. To further improve glyph fidelity, we propose region-level human preference optimization. We construct the region-level preference dataset \datasetname and design Region-Grouped DPO (R-GDPO) to optimize the model with region-level preference pairs instead of image-level ones. By explicitly aligning model outputs with accurate regions while discouraging incorrect ones, R-GDPO enables more precise control over glyph quality and significantly improves glyph accuracy.

\myparagraph{Construction of \datasetname Dataset.}
Inspired by how humans correct spelling errors through feedback on glyph mistakes, we construct \datasetname with fine-grained annotations of correct and incorrect glyph regions. \textbf{1).} We sample prompt-glyph pairs $\big(\mathcal{P},\left\{(\mathcal{T}_i,\mathbf{P}_i)\right\}_{i=1}^{N_t}\big)$  from the collected realistic data in Stage 1. \textbf{2).} We replace each text condition $\mathcal{T}_i$ with characters from a pool containing English letters and Chinese characters, leading to the new glyph condition $\left\{(\mathcal{T}_i^{\prime},\mathbf{P}_i)\right\}_{i=1}^{N_t}$. \textbf{3).} For each new prompt–glyph pair, we use the Stage 1 model to generate a group of candidate images. Human annotators then label the correct and incorrect text regions in each image, used for constructing region-level preference pairs that capture localized glyph correctness.
For each prompt–glyph pair, we sample 10 images and filter out those with low overall image aesthetics. In total, we obtain 7{,}117 images from 879 prompt–glyph pairs. An example from \datasetname is shown in \cref{fig:pipeline}~(b), where incorrect glyphs with extraneous or missing strokes are highlighted with green boxes.

\myparagraph{Region-Grouped Direct Preference Optimization.} Standard DPO~\cite{wallace2024diffusion} leverages preference image pairs to align the model's output with winning samples while distancing it from losing ones. However, it is unsuitable for text rendering since incorrect glyphs often appear in localized regions, as shown in \cref{fig:pipeline} (b). Therefore, we introduce \dpomethodname, leveraging the region-level preference pairs derived from \datasetname. First, we conceptualize the denoising process as a multi-step Markov Decision Process, and modify the RL objective to its region-level variant:
\begin{equation}
\begin{aligned}\label{eq:mdpo_objective}
\underset{p_\theta}{\text{min}} - &E_{c,x_0\sim p_\theta(x_0|c)}[r_m(x_{0},c,M)/\beta] + \Sigma_{t=1}^{T}E_{c,p_\theta(x_t|c)}[ \\
&\mathcal{D}_\text{KL}\big(p_\theta(M(x_{t-1})|x_t,c) || p_\text{ref}(M(x_{t-1})|x_t,c)\big)],\\ 
\end{aligned}
\end{equation}
where we use the model from Stage 1 as $p_\text{ref}$, and $M(\cdot)$ selects the elements within the given mask $M$. Then, the regional reward function $r_m$ can be derived as:
\begin{multline}\label{eq:mdpo_reward}
r_m(x_{0},c,M)=\beta T E_{t,x_{t-1,t}\sim p^*_\theta(x_{t-1,t}|x_0,c)}\Big[\\ \text{log}\frac{p_\theta^*(M(x_{t-1})|x_{t},c)}{p_\text{ref}(M(x_{t-1})|x_{t},c)} \Big]+Z_m(c,M).
\end{multline} 
Essentially, $r_m$ evaluates the given region in the image under the condition $c$. The detailed derivation and the form of $Z_m(c,M)$ are presented in the supplementary. The training objective is then given by:
\begin{multline}\label{eq:mdpo_loss}
L_\text{R-DPO}=-E_{c,x_0^w,x_0^l,M^w,M^l}\Big[\text{log}\sigma\big(r_m(x_0^w,c,M^w) \\
-r_m(x_0^l,c,M^l) \big)\Big]\\
\leq-E_{c,t,x_0^w,x_0^l,M^w,M^l,\epsilon^w,\epsilon^l}\Big[\text{log}\sigma\Big(-\beta T \omega_t\big(\\
||M^w(v^w-v_\theta(x^w_t,t,c))||^2_2- ||M^w(v^w-v_\text{ref}(x^w_t,t,c))||^2_2 \\
-(||M^l(v^l-v_\theta(x^l_t,t,c))||^2_2 -||M^l(v^l-v_\text{ref}(x^l_t,t,c))||^2_2)\big) \\+\Delta Z_m(c,M^w,M^l) \Big)\Big], \\
\end{multline}
where $M^w$ and $M^l$ indicate the winning and losing regions from $x_0^w$ and $x_0^l$, respectively, and $\Delta Z_m(c,M^w,M^l)$ takes the form $Z_m(c,M^w)-Z_m(c,M^l)$. Compared with standard DPO loss in \cref{eq:dpo_loss}, \cref{eq:mdpo_loss} considers the region-level preference pairs, which may come from the same image, \ie $x^w=x^l$. It's worth noting that $Z_m$ terms are not canceled in the \cref{eq:mdpo_loss} when $M^w \neq M^l$, since $\Delta Z_m(c,M^w,M^l) \neq 0$. The derivation details of \cref{eq:mdpo_loss} are provided in the supplementary.

Next, we need to construct region-level winning-losing pairs for each group from \datasetname. To make more efficient use of data, we consider the following masks for a sample pair $(i,j)$ within a group. First, $M_{i,j}^{+,-}$ indicates the region where sample $i$ is correct but sample $j$ is incorrect. In addition, $M_{i}^{+}$/$M_{i}^{-}$ denotes the correct/incorrect region in sample $i$. Specifically, $M_{i,j}^{+,-}$ selects the winning-losing pairs across different images, termed \textit{inter-sample preference mask}. In contrast, $M_{i}^{+}$ and $M_{i}^{-}$ identify the preference pairs within the same sample, termed \textit{intra-sample preference masks}. \cref{fig:pipeline} (b) shows an example of the above masks. Then, we consider all preference pairs within a group and extend the objective from \cref{eq:mdpo_loss} to:
\begin{equation}
\begin{aligned}\label{eq:rmdpo_loss}
&L_\text{R-GDPO}=-E\Big[\frac{1}{N_G}\Sigma_{i=1}^{N_G}\Sigma_{j=1}^{N_G} \lambda_{i,j}\text{log}\sigma\big(-\beta T\omega_tL_{i,j} \big)\Big], \\ 
&\lambda_{i,j}, L_{i,j} =
\begin{cases}
  {\lambda_{\text{inter}}}/({N_G - 1}), L_{\text{inter}}, & i \neq j, \\
  % [6pt]
   1 - \lambda_{\text{inter}}, L_{\text{intra}}, & i = j, \\
\end{cases} \\
\end{aligned}
\end{equation}
where $N_G$ is the group size. $\lambda_\text{inter}$ controls the weights of $L_\text{inter}$ and $L_\text{intra}$, which use \textit{inter-sample} and \textit{intra-sample preference masks}, respectively:
\begin{equation}
\resizebox{0.48\textwidth}{!}{$
\begin{aligned}\label{eq:rmdpo_loss_2}
&L_\text{inter}= ||M_{i,j}^{+,-}(v^i-v_\theta(x^i_{t},t,c))||^2_2-||M_{i,j}^{+,-}(v^i-v_\text{ref}(x^i_t,t,c))||^2_2 \\
&-\big(||M_{i,j}^{+,-}(v^j-v_\theta(x^j_t,t,c))||^2_2-||M_{i,j}^{+,-}(v^j-v_\text{ref}(x^j_t,t,c))||^2_2\big), \\
&L_\text{intra}=||M_{i}^{+}(v^i-v_\theta(x^i_{t},t,c))||^2_2-||M_{i}^{+}(v^i-v_\text{ref}(x^i_t,t,c))||^2_2 \\
&-\big(||M_{i}^{-}(v^i-v_\theta(x^i_t,t,c))||^2_2-||M_{i}^{-}(v^i-v_\text{ref}(x^i_t,t,c))||^2_2\big), \\
\end{aligned}
$}
\end{equation}
where $\Delta Z_m$ terms are discarded for $L_{i,j}$ by setting an appropriate $\lambda_{\text{inter}}$. Specifically, we assign a high value to $\lambda_\text{inter}$ since $\Delta Z_m$ is negligible in $L_\text{inter}$, given that $M^w=M^l$. \cref{eq:rmdpo_loss} is the proposed \dpomethodname objective.

\begin{figure*}[t]
    \centering
    \includegraphics[width=1\linewidth]{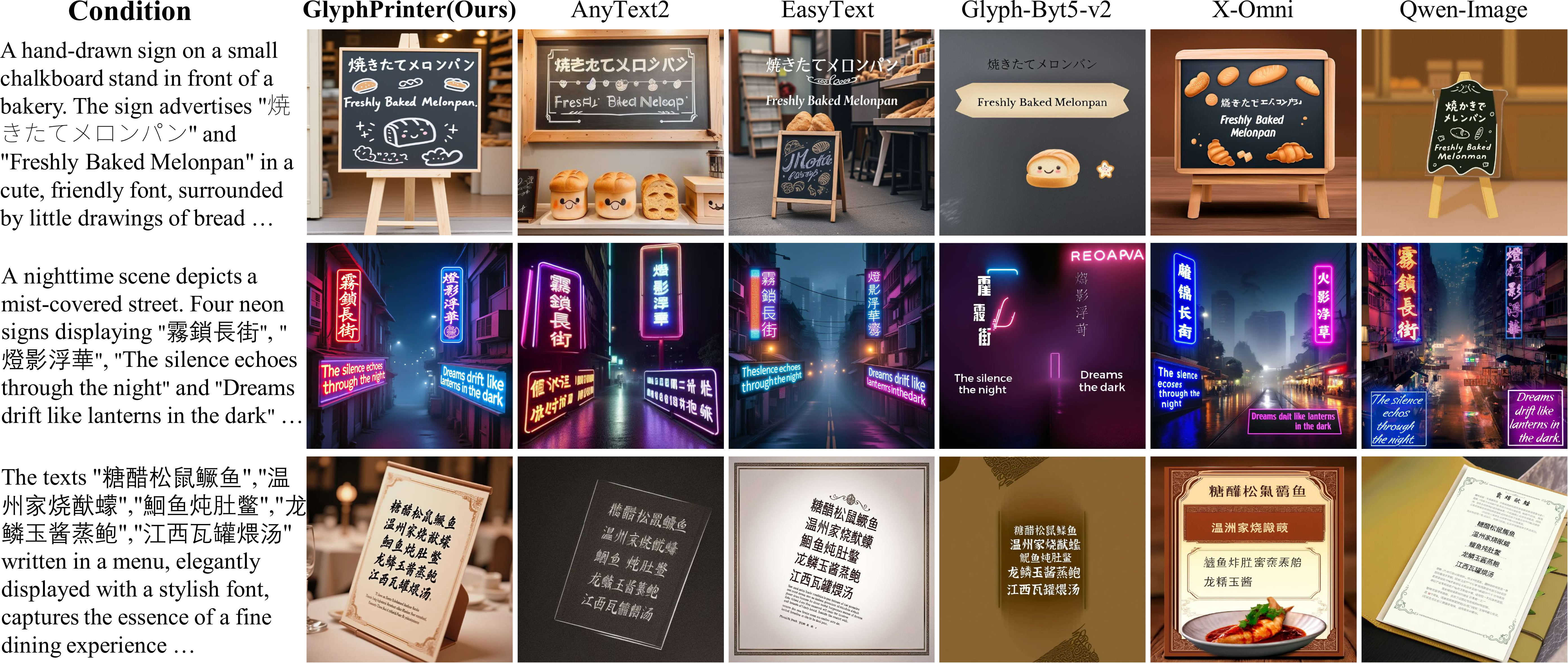}
    \vspace{-6mm}
    \caption{Text rendering results under simple and complex glyph conditions. Our \methodname outperforms other methods in preserving fine details of glyphs, particularly for some complex cases in the 2nd and 3rd rows. Best viewed zoomed in to see the fine glyph details.}
    \label{fig:exp1}

\end{figure*}

\subsection{Regional Reward Guidance}
Through the two-stage training, our text rendering model can already generate accurate glyphs with diverse styles. To further provide more controllability during inference, we introduce
\completeinfermethodname (\infermethodname), inspired by Classifier-Free Guidance~\cite{ho2022classifier}. Concretely, the models from Stage 1 and Stage 2 approximate $p( x_t| c)$ and $p( x_t| c,r)$, where $r$ can be regarded as a ``reward condition'' that increases the glyph accuracy. In this setting, we can replace $p(x_t)$ and $p(x_t|\mathcal{C})$ from \cref{eq:cfg} with $p_\text{ref}(x_t|c)$ and $p_\theta(x_t|c)$:
\begin{equation}
\begin{aligned}\label{eq:dpo_cfg}
\hat{s}_{\text{ref},\theta}(x_t,\omega,c)&=\nabla_{x_t}\text{log}\big(p_\text{ref}(x_t|c)^{(1-\omega)}p_\theta(x_t|c)^\omega\big). \\
\end{aligned}
\end{equation}
As derived in the supplementary, \cref{eq:dpo_cfg} yields the optimal distribution $p_\text{ref}(x_0|c)\text{exp}(\frac{r(x_0,c)}{\beta/w})$, where $\beta/w$ is an adjustable regularization weight that controls glyph accuracy during inference. Since $v(x_t,t)=a_tx+b_t \nabla_{x_t}\text{log}p(x_t)$ \cite{zheng2023guided}, where the parameters $a_t$ and $b_t$ are determined by the scheduler, we can derive the corresponding velocity field for this optimal distribution:
\begin{equation}
\begin{aligned}\label{eq:dpo_cfg_2}
v^*(x_t,t,c)&= a_tx+b_t \big(\nabla_{x_t}\text{log}\big(p_\text{ref}(x_t|c)^{(1-\omega)}p_\theta(x_t|c)^\omega\big)\big)\\
&= (1-\omega)v_\text{ref}(x_t,t,c)+\omega v_\theta(x_t,t,c).\\
\end{aligned}
\end{equation}
The derivation details are presented in the supplementary. The proposed \infermethodname is then defined as:
\begin{equation}
\begin{aligned}\label{eq:mdpo_cfg}
\hat{v}^*(x_t,t,c)&=\hat{\mathbf{P}}{v}^*(x_t,t,c)+(\textbf{I}-\hat{\mathbf{P}}){v}_\theta(x_t,t,c),\\
\end{aligned}
\end{equation}
where $\hat{\mathbf{P}}$ indicates the overall text region derived from the given positions $\{\mathbf{P}_i\}_{i=1}^{N_t}$. As indicated in \cref{eq:mdpo_cfg}, \infermethodname maintains the image quality in the background.

\section{Experiments}
\subsection{Experimental Setup}
\myparagraph{Implementation Details.}~The proposed \methodname builds upon Flux.1-Dev~\cite{black2024flux}. All experiments are conducted using 4 NVIDIA A800 80G GPUs. In {Stage~1}, synthetic data are created by overlaying multilingual characters on images from the LAION dataset, while realistic data are collected from the internet. Following~\cite{lu2025easytext}, we first learn the injected LoRAs on the synthetic data for 30{,}000 steps at a resolution of $512 \times 512$, followed by 10{,}000 steps at $1024 \times 1024$. Then, we introduce additional LoRAs and optimize them on the realistic part for 15{,}000 steps at $1024 \times 1024$. The batch size is set to 16 throughout Stage 1. In {Stage~2}, we leverage the constructed \datasetname and optimize another set of LoRAs for 4{,}000 steps with a batch size of 16 and a group size of 4. $\beta$ is set to 2 and $\lambda_\text{inter}$ is set to 0.7. The rank numbers of introduced LoRAs are all set to 128.

\myparagraph{Evaluation Details.} 
We use Vision Language Model \cite{wang2024qwen2} to evaluate the following metrics: \textbf{1)} Image-text alignment (\textbf{Text.Align}). \textbf{2)} Image aesthetic (\textbf{Aes}). \textbf{3)} Normalized Edit Distance (\textbf{NED}) and sentence accuracy (\textbf{Sen.Acc}) for measuring text accuracy. Furthermore, since existing text recognition models are insensitive to incorrect glyphs, we conduct a user study as a complementary evaluation, where we introduce a new metric \textbf{Glyph.Acc} for the accuracy of the generated glyphs. To assess model performance across various scenarios, we construct two benchmarks: \multilingualbenchmarkname and \complexbenchmarkname, including multilingual and complex glyph conditions, respectively. 

For compared methods, the {prompt-guided} approaches include Glyph-ByT5-v2~\cite{liu2024glyphv2}, X-Omni~\cite{geng2025x} and Qwen-Image~\cite{wu2025qwen}. The selected {glyph-image-guided} methods are AnyText2~\cite{tuo2024anytext2} and EasyText~\cite{lu2025easytext}. More details and results of the evaluation can be found in the supplementary.

\begin{figure*}[t]
    \centering
    \includegraphics[width=1\linewidth]{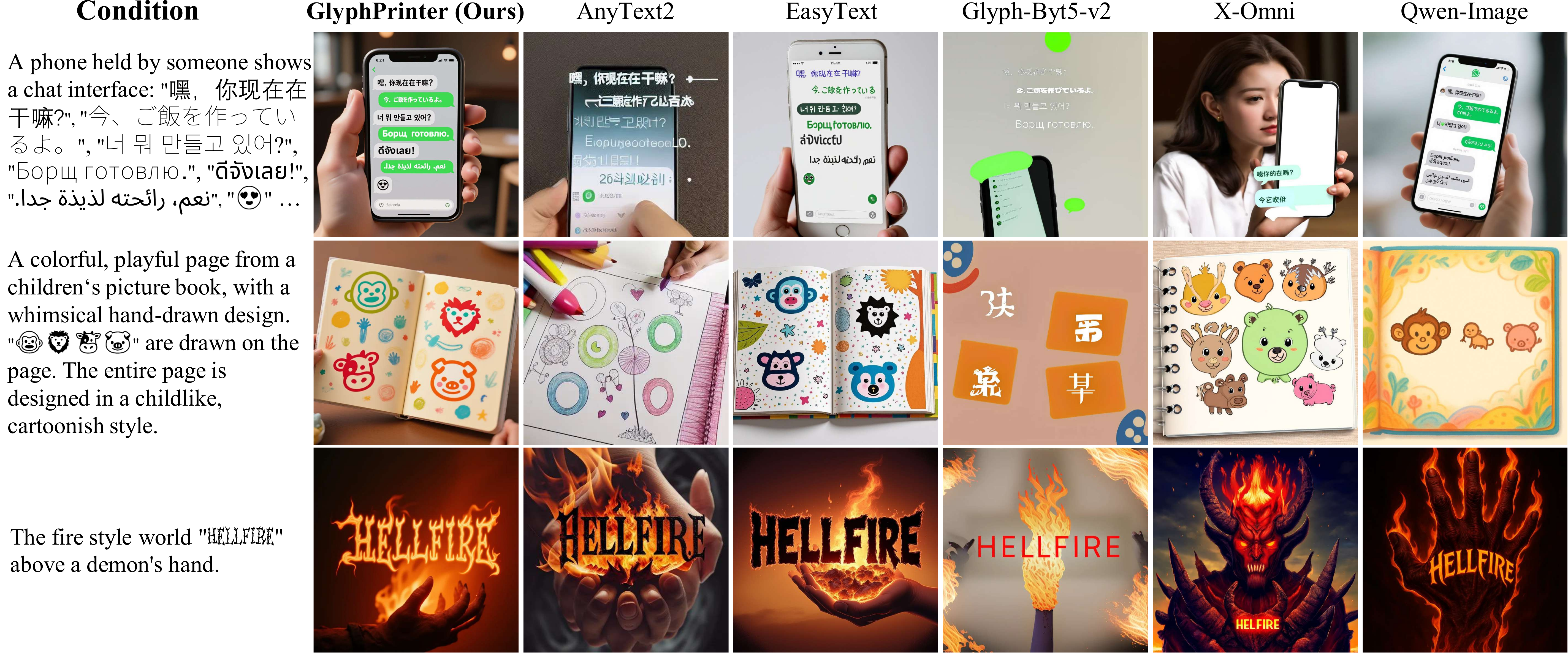}
    \vspace{-6mm}
    \caption{Text rendering results under multilingual and out-of-domain glyph conditions. Our method \methodname consistently preserves glyph fidelity, indicating strong generalization ability. Please zoom in for better visibility of fine text details.}
    \label{fig:exp2}

\end{figure*}

\begin{table*}[t]
\centering
\footnotesize
\caption{Evaluation of text rendering accuracy. The rows from ``English'' to ``Thai' indicate the results of \multilingualbenchmarkname, while the ``Complex'' row shows the result of \complexbenchmarkname. Bold indicates the best result, and underline denotes the second best.}
\vspace{-2mm}
\label{tab:benchmark_comparison}
% \scriptsize
\setlength{\tabcolsep}{3.6pt}
\renewcommand{\arraystretch}{1.1}
\resizebox{\textwidth}{!}{ 
\begin{tabular}{c|cc|cc|cc|cc|cc|cc}
\toprule
\textbf{Method} & \multicolumn{2}{c|}{\textbf{\methodname (Ours)}} & \multicolumn{2}{c|}{AnyText2~\cite{tuo2024anytext2}} & \multicolumn{2}{c|}{EasyText~\cite{lu2025easytext}} & \multicolumn{2}{c|}{Glyph-Byt5-v2~\cite{liu2024glyphv2}} & \multicolumn{2}{c|}{X-Omni~\cite{geng2025x}} & \multicolumn{2}{c}{Qwen-Image~\cite{wu2025qwen}} \\
\cline{2-13}
 & \textbf{NED} & \textbf{Sen.Acc} & \textbf{NED} & \textbf{Sen.Acc} & \textbf{NED} & \textbf{Sen.Acc} & \textbf{NED} & \textbf{Sen.Acc} & \textbf{NED} & \textbf{Sen.Acc} & \textbf{NED} & \textbf{Sen.Acc} \\
\hline
English & \textbf{0.9851} & \textbf{0.9320} & 0.8162 & 0.5291 & 0.9780 & 0.9174 & \underline{0.9844} & \underline{0.9223} & 0.7963 & 0.5146 & 0.9193 & 0.8689 \\
Chinese & \textbf{0.9728} & \textbf{0.9339} & 0.9004 & 0.7899 & \underline{0.9569} & 0.9144 & 0.9295 & 0.8093 & 0.7318 & 0.4397 & 0.9149 & \underline{0.9183} \\
Japanese & \textbf{0.9616} & \textbf{0.9167} & 0.8543 & 0.6111 & \underline{0.9389} & \underline{0.8888} & 0.9038 & 0.8056 & 0.5148 & 0.2222 & 0.7324 & 0.5278 \\
Korean & \textbf{0.9693} & \textbf{0.9362} & 0.7222 & 0.4894 & 0.8544 & 0.7660 & \underline{0.9648} & \underline{0.8511} & 0.0670 & 0.0213 & 0.5167 & 0.2128 \\
French & \underline{0.9714} & 0.8427 & 0.7321 & 0.3270 & 0.9671 & \underline{0.8553} & \textbf{0.9740} & \textbf{0.8742} & 0.6939 & 0.2390 & 0.7470 & 0.5220 \\
Vietnamese & \textbf{0.9742} & \textbf{0.8500} & 0.5714 & 0.1000 & \underline{0.9706} & \underline{0.8250} & 0.8739 & 0.4500 & 0.5538 & 0.1000 & 0.6209 & 0.2000 \\
Thai & \textbf{0.9663} & \textbf{0.8413} & 0.0431 & 0.0159 & \underline{0.9166} & \underline{0.8095} & 0.0040 & 0.0000 & 0.0368 & 0.0000 & 0.2523 & 0.0476 \\
\hline
Complex & \textbf{0.9013} & \textbf{0.8349} & \underline{0.7867} & 0.6368 & 0.7645 & \underline{0.6604} & 0.6179 & 0.3396 & 0.2972 & 0.0755 & 0.6189 & 0.3679 \\
\bottomrule
\end{tabular}
}
\end{table*}

\begin{table}[t!]
\centering
\small
\setlength{\tabcolsep}{5pt}
\caption{The user study assesses image aesthetic (\textbf{Aes}), text-image alignment (\textbf{Text.Align}),  and glyph accuracy (\textbf{Glyph.Acc}). All scores are within the range of 1 to 10.}
\vspace{-2mm}

\renewcommand{\arraystretch}{1.1}

\begin{tabular}{r|ccc}
\toprule
\textbf{Method} & \textbf{Aes} & \textbf{Text.Align} & \textbf{Glyph.Acc} \\
\hline
\textbf{\methodname} (\textbf{Ours}) & 7.9143 & \underline{8.1692} & \textbf{8.3084} \\
\methodname-Stage~1 & \underline{7.9507} & 8.0899 & 7.5332\\
Anytext2~\cite{tuo2024anytext2} & 5.5310 & 6.3298 & 5.3769 \\
EasyText~\cite{lu2025easytext} & 7.0964 & 7.6767 & 7.4411 \\
Glyph-Byt5-v2~\cite{liu2024glyphv2} & 3.4454 & 3.7173 & 6.3919 \\
X-Omni~\cite{geng2025x} & 6.6660 & 7.1328 & 5.8972 \\
Qwen-Image~\cite{wu2025qwen} & \textbf{8.0749} & \textbf{8.2291} & \underline{7.9657} \\
\bottomrule
\end{tabular} \label{tab:user_study_transposed}
\vspace{-0.3cm}
\end{table}

\begin{table*}[t]
\centering
\caption{Evaluation of text-image alignment and image aesthetic. Both scores are within the range of 1 to 100.}
\vspace{-2mm}
\label{tab:aes_alignment}
\scriptsize
\setlength{\tabcolsep}{1pt}
\renewcommand{\arraystretch}{1.1}
\resizebox{\textwidth}{!}{ % Resizes the table to fit the text width
\begin{tabular}{c|cc|cc|cc|cc|cc|cc}
\toprule
\textbf{Method} & \multicolumn{2}{c|}{\textbf{\methodname (Ours)}} & \multicolumn{2}{c|}{AnyText2~\cite{tuo2024anytext2}} & \multicolumn{2}{c|}{EasyText~\cite{lu2025easytext}} & \multicolumn{2}{c|}{Glyph-Byt5-v2~\cite{liu2024glyphv2}} & \multicolumn{2}{c|}{X-Omni~\cite{geng2025x}} & \multicolumn{2}{c}{Qwen-Image~\cite{wu2025qwen}} \\
\cline{2-13}
 & \textbf{Text.Align} & \textbf{Aes} & \textbf{Text.Align} & \textbf{Aes} & \textbf{Text.Align} & \textbf{Aes} & \textbf{Text.Align} & \textbf{Aes} & \textbf{Text.Align} & \textbf{Aes} & \textbf{Text.Align} & \textbf{Aes} \\
\hline
Multilingual & \underline{84.2992} & 87.3315 & 69.1752 & 82.5957 & 83.2642 & 86.5148 & 72.7898 & 82.5067 & 79.0825 & \textbf{89.6392} & \textbf{88.1968} & \underline{87.8167} \\
Complex & \underline{91.8969} & \textbf{89.5330} & 84.4742 & 86.2886 & 88.2371 & 89.2268 & 81.7010 & 84.1237 & 81.8652 & 87.4663 & \textbf{92.0825} & \underline{89.4845} \\
\bottomrule
\end{tabular}
}
\end{table*}

\begin{table*}[t]
\centering
\caption{Ablation study. Bold indicates the best result, and underline denotes the second best.}
\label{tab:ablation}
\vspace{-2mm}
\scriptsize
\setlength{\tabcolsep}{2pt}
\renewcommand{\arraystretch}{1.1}
\resizebox{\textwidth}{!}{ % Resizes the table to fit the text width
\begin{tabular}{c|cc|cc|cc|cc|cc|cc|cc}
\toprule
\textbf{Method} & \multicolumn{2}{c|}{\textbf{\methodname (Ours)}} & \multicolumn{2}{c|}{{Stage 1}} & \multicolumn{2}{c|}{SFT} & \multicolumn{2}{c|}{Mask SFT} & \multicolumn{2}{c|}{{w/o RRG}} & \multicolumn{2}{c|}{$\lambda_\text{inter}=1$}& \multicolumn{2}{c}{$\lambda_\text{inter}=0$}   \\
\cline{2-15}
& \textbf{NED} & \textbf{Sen.Acc} & \textbf{NED} & \textbf{Sen.Acc} & \textbf{NED} & \textbf{Sen.Acc} & \textbf{NED} & \textbf{Sen.Acc} & \textbf{NED} & \textbf{Sen.Acc} & \textbf{NED} & \textbf{Sen.Acc} & \textbf{NED} & \textbf{Sen.Acc} \\
\hline
Multilingual & \textbf{0.9715} & \textbf{0.8932} & 0.9360 & 0.8023 & 0.9357 & 0.8238 & 0.9483 & 0.8275  & 0.9522 & \underline{0.8772} & \underline{0.9605} & 0.8725 &  0.9549 & 0.8458\\
Complex & \textbf{0.9013} & \textbf{0.8349} & 0.8142 & 0.6981 & 0.8418 & 0.7406 & 0.8437 & 0.7547 & 0.8810 & 0.8019 & \underline{0.8908} & \underline{0.8066} &  0.8781 & 0.7925\\
\bottomrule
\end{tabular}
}
\end{table*}

\subsection{Performance Comparison on Text Rendering}
Herein, we evaluate the model's rendering accuracy under various scenarios. We first provide some examples to illustrate the model's performance for both simple and complex glyph conditions. As shown in the 1st row of \cref{fig:exp1}, most of the methods can generate accurate characters with simple glyphs. However, AnyText2~\cite{tuo2024anytext2} exhibits inferior accuracy due to the excessive stylization. In contrast, Glyph-ByT5-v2~\cite{liu2024glyphv2} tends to produce rigid visual styles. Furthermore, the compared methods struggle to create complex glyphs. As shown in the 2nd and 3rd rows of \cref{fig:exp1}, even advanced methods trained on large-scale scene text images, such as Qwen-Image~\cite{wu2025qwen} and X-omni~\cite{geng2025x}, fail to generate accurate glyphs for complex Chinese characters. Without RL post-training, EasyText~\cite{lu2025easytext} is prone to generating glyphs with extraneous or missing strokes. In comparison, our proposed \methodname exhibits better accuracy in all cases.

We also construct several challenging glyph conditions containing multilingual texts, emojis, and characters with stylized fonts, to further assess the generalization ability. As shown in the 1st and 2nd rows of \cref{fig:exp2}, the results from {prompt-guided} methods, such as X-Omni and Qwen-Image, often omit some characters due to the limited vocabulary. Furthermore, they fail to generate desired results when conditioned on unseen font names, as shown in the 3rd row of \cref{fig:exp2}. In contrast, the outcomes of {glyph-image-guided} approaches~\cite{tuo2024anytext2,lu2025easytext} exhibit better alignment with glyph conditions. However, they struggle to faithfully capture the structural details. As shown in the 2nd and 3rd rows of \cref{fig:exp2}, the emojis and stylized text generated by EasyText lose the fine details presented in the input condition.

Quantitative results in \cref{tab:benchmark_comparison} also demonstrate that the proposed \methodname outperforms in text accuracy for both \multilingualbenchmarkname and \complexbenchmarkname. Moreover, the {Glyph.~Acc} metric from \cref{tab:user_study_transposed} indicates that our method generates more accurate glyphs, while other methods are more prone to generating glyphs with extraneous or missing strokes. In addition, the text-image alignment and image aesthetic metrics presented in \cref{tab:aes_alignment} and \cref{tab:user_study_transposed} show that our approach achieves comparable results with the method~\cite{lu2025easytext} building upon the same base model~\cite{black2024flux}. However, due to the differences between fundamental models, some of these metrics are slightly inferior to X-Omni and Qwen-Image. 

\begin{figure}[t]
    \centering
    % \vspace{-3.0mm}
    \includegraphics[width=0.476\textwidth]{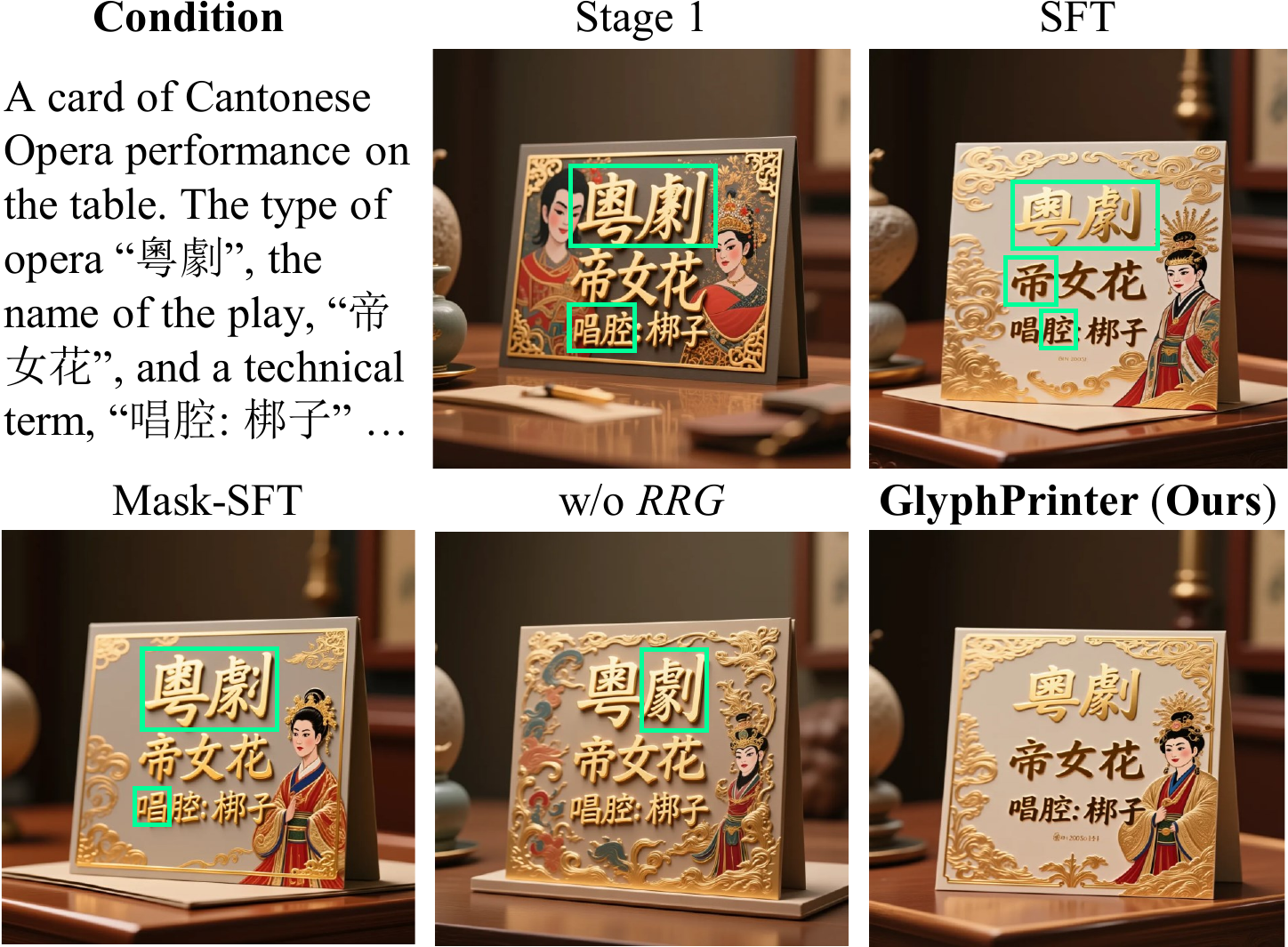}
    \vspace{-5.6mm}
    \caption{Ablation studies. The erroneous regions of each image are highlighted with green boxes.}
    \label{fig:ablation}
    % \vspace{-1mm}
\end{figure}

\subsection{Ablation Studies}
We conduct ablation studies to analyze the impacts of our key designs. First, we demonstrate the effect of the proposed \dpomethodname and compare our method with the following settings: Stage 1, SFT (the loss in \cref{eq:flow_loss} is applied in Stage 2), and Mask-SFT (the loss in \cref{eq:flow_loss} is applied only to the correct text regions in Stage 2). \cref{fig:ablation} presents qualitative results. As indicated in the figure, the model from Stage 1 exhibits suboptimal performance due to the limited glyph distribution in the training data. Without leveraging region-level annotations, the model with SFT performs comparably to the Stage 1 setting. In contrast, the model with Mask-SFT achieves better results than the former settings. However, due to the lack of penalties for incorrect glyphs, the model is still prone to generating characters with extraneous or missing strokes. Furthermore, the result of the model without \infermethodname shows subtle flaws in some small areas. In contrast, our method outperforms these settings. The quantitative results in \cref{tab:ablation} also demonstrate the effects of our designs. Additionally, we train the models with only $L_\text{inter}$ and $L_\text{intra}$, corresponding to $\lambda_\text{inter}=1$ and $\lambda_\text{inter}=0$ in \cref{tab:ablation}, respectively. Due to inefficient usage of region-level preference samples, the model trained with $L_\text{inter}$ performs worse than our method. The model using $L_\text{intra}$ also exhibits lower performance since $\Delta Z_m$ cannot be neglected in this setting, leading to an imprecise training objective.

\section{Conclusion}
We propose a preference-based text rendering framework, \methodname, to address the challenge of generating accurate glyphs for visual text rendering. We first construct the dataset \datasetname with region-level preference annotations. Building on this dataset, we introduce Region-Grouped DPO to optimize the model with region-level preference pairs, which better aligns model outputs with correct glyphs while discouraging incorrect ones. In addition, the proposed Regional Reward Guidance further enhances performance at inference time by providing controllable glyph accuracy. Extensive experiments demonstrate that the proposed approach outperforms existing text rendering methods in glyph accuracy while maintaining a favorable balance between stylization and precision.\\
\textbf{Limitations.} Although \methodname achieves high glyph accuracy for text rendering, it is less accurate when rendering very small characters due to the limitations of the VAE. Moreover, it is trained on images with a fixed resolution, which limits its ability to produce outputs with diverse aspect ratios. We leave these issues for future work.

% \clearpage

% \clearpage
% \twocolumn
{
    \small
    \bibliographystyle{ieeenat_fullname}
    \bibliography{main}
}

{
\normalsize
\clearpage
\onecolumn

\setcounter{section}{0}
\setcounter{figure}{0}
\setcounter{table}{0}
\renewcommand{\thesection}{\Alph{section}}
\renewcommand{\thetable}{\Roman{table}}
\renewcommand{\thefigure}{\Roman{figure}}
\renewcommand{\theequation}{\roman{equation}}

\begin{center}
\vspace{1.6em}
\textbf{\Large{Supplementary Material for \methodname}}\\
\vspace{3.6em}
\end{center}

\section{More Details of Stage 1}
\subsection{Construction of the Training Dataset}
Following previous works~\cite{liu2024glyphv2,wu2025qwen,lu2025easytext}, we train the \methodname with both a synthetic dataset and a collected realistic dataset. The synthetic dataset is used to improve the model's text rendering ability. Specifically, we use 560K images from LAION dataset as background, which are then overlaid with multilingual characters (languages include Latin, Chinese, Japanese, Korean, Arabic, and Thai) through the following steps: \textbf{1)} For each background image, we first use an optimization algorithm to obtain 1 to 4 non-overlapping bounding boxes, each with a random orientation, rotation angle, and size. \textbf{2)} For each bounding box, we randomly select a language and insert an appropriate number of characters based on the box size. \textbf{3)} We randomly select bounding boxes and apply distortion to them. Several samples are provided on the left of~\cref{fig:appendix-dataset}. 

To generate text images with diverse styles, we additionally collect 30K high-quality text images from the internet. These images are then annotated with an OCR model~\cite{cui2025paddleocr} and VLM~\cite{bai2025qwen2} for text recognition and image caption, respectively. Following the previous work~\cite{lu2025easytext}, placeholders such as \texttt{<sks1>}, \texttt{<sks2>}, \ldots, \texttt{<sksn>} are used in captions to represent the text from each text block. This dataset mainly contains Chinese and English texts, as illustrated on the right side of \cref{fig:appendix-dataset}.

\begin{figure*}[htbp]
    \centering
    \includegraphics[width=1\textwidth]{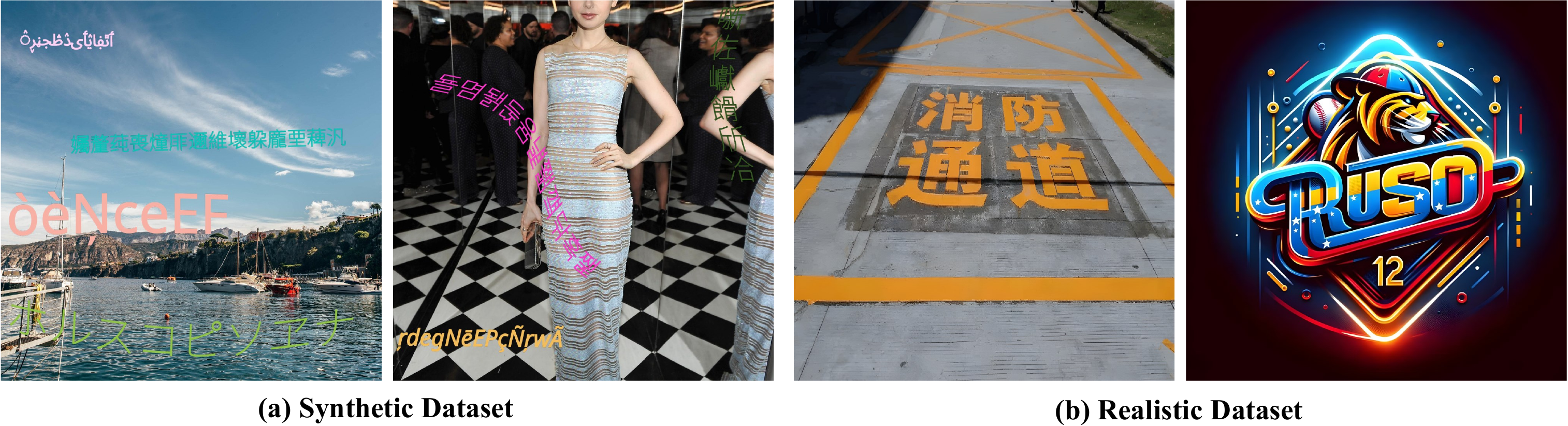}
    \caption{Samples from the \textbf{(a)} synthetic dataset and \textbf{(b)} realistic dataset used in Stage 1.}
    \label{fig:appendix-dataset}
\end{figure*}

\begin{figure*}[htbp]
    \centering
    \includegraphics[width=1\textwidth]{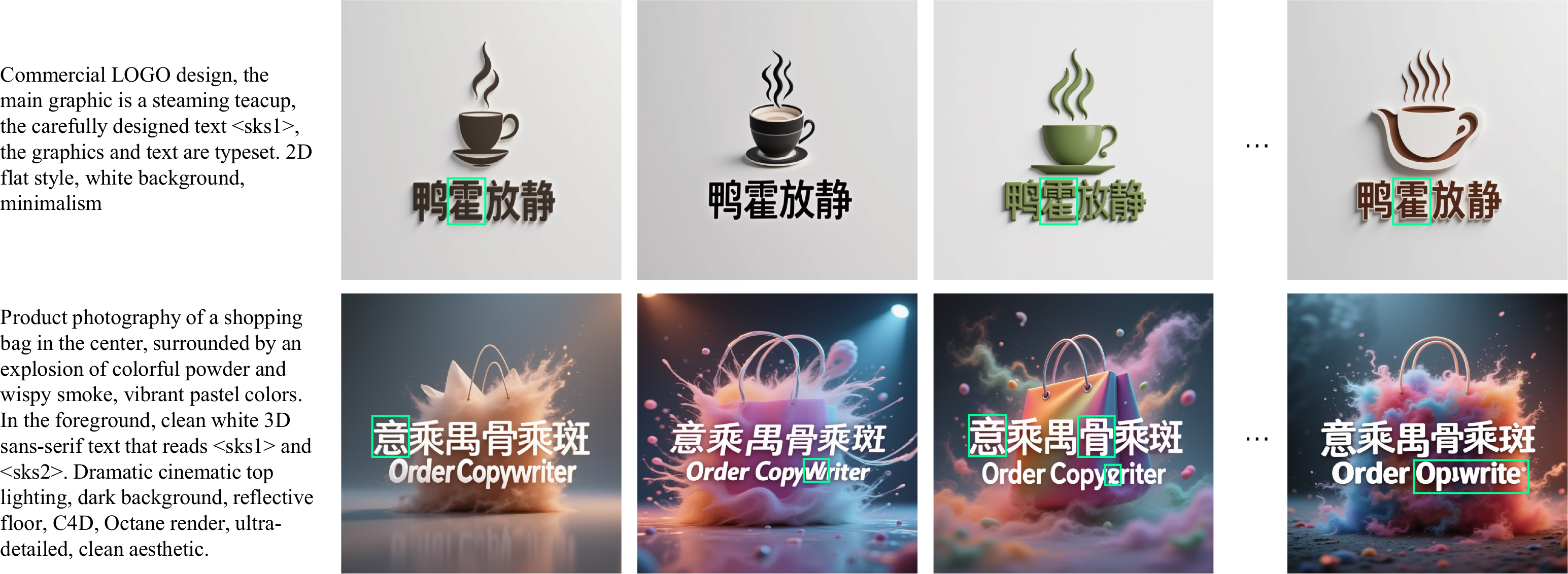}
    \caption{Samples from the proposed \datasetname. The erroneous regions are highlighted with green boxes.}
    \label{fig:appendix-glyphcorrector}
\end{figure*}

\section{More Details of Stage 2}
\subsection{Construction of \datasetname} 
In Stage 2, we construct \datasetname to make the model learn from region-level preference pairs, improving the glyph accuracy. Each annotator labels their assigned samples by comparing generated characters with the synthetic rendered ones, followed by cross-review and iterative correction until all errors are resolved. Some examples are shown in \cref{fig:appendix-glyphcorrector}.

\paragraph{The motivation for generating a group of images per condition.} The traditional DPO objective~\cite{rafailov2023direct} only considers a single preference pair in each training batch, which is insufficient for visual text rendering tasks. This is because a small number of samples are difficult to cover completely accurate glyphs, particularly for long or complex glyph conditions. For instance, given the glyph condition ``12345678'', the first sample contains accurate ``1234'', while the second one accurately renders ``3456''. However, the model still fails to learn the accurate ``78'' with this single preference pair. Therefore, generating a group of images for each condition enhances sample diversity, enabling the model to learn accurate glyphs from various samples and thereby improving model performance.

\subsection{Derivation of Eq. (8)} \label{sec:appendix_derivation_8}
Here, we derive that Eq. (8) is exactly the implicit reward model learned from Eq. (7). First, we reformulate Eq. (7) as:
\begin{equation}
\begin{aligned}\label{eq:mdpo_objective_2} 
&\underset{p_\theta}{\text{min}} - E_{c,x_0\sim p_\theta(x_0|c)}[r_m(x_{0},c,M)/\beta] + \Sigma_{t=1}^{T}E_{c,p_\theta(x_t|c)}[\mathcal{D}_\text{KL}\big(p_\theta(M(x_{t-1})|x_t,c) || p_\text{ref}(M(x_{t-1})|x_t,c)\big)]\\ 
=&\underset{p_\theta}{\text{min}} - E_{c,p_\theta(x_{0:T}|c)}[\Sigma_{t=0}^{T-1} R_m(M(x_{t}),c)/\beta] + \Sigma_{t=1}^{T}E_{c,p_\theta(x_t|c)}[\mathcal{D}_\text{KL}\big(p_\theta(M(x_{t-1})|x_t,c) || p_\text{ref}(M(x_{t-1})|x_t,c)\big)]\\ 
= &\underset{p_\theta}{\text{min}} - E_{c,p_\theta(x_{0:T}|c)}[ \Sigma_{t=0}^{T-1} R_m(M(x_{t}),c)/\beta] + \Sigma_{t=1}^{T}E_{c,p_\theta(x_t|c)}E_{p_\theta(M(x_{t-1})|x_t,c)}\Big[\text{log}\frac{p_\theta(M(x_{t-1})|x_t,c)}{p_\text{ref}(M(x_{t-1})|x_t,c)}\Big]\\ 
= &\underset{p_\theta}{\text{min}} - E_{c,p_\theta(x_{0:T}|c)}[ \Sigma_{t=0}^{T-1} R_m(M(x_{t}),c)/\beta] + \Sigma_{t=1}^{T}E_{c,p_\theta(x_t|c)}E_{p_\theta(x_{t-1}|x_t,c)}\Big[\text{log}\frac{p_\theta(M(x_{t-1})|x_t,c)}{p_\text{ref}(M(x_{t-1})|x_t,c)}\Big]\\ 
= &\underset{p_\theta}{\text{min}} - E_{c,p_\theta(x_{0:T}|c)}[ \Sigma_{t=0}^{T-1} R_m(M(x_{t}),c)/\beta] + E_{c,p_\theta(x_{0:T}|c)}\Big[\Sigma_{t=1}^{T}\text{log}\frac{p_\theta(M(x_{t-1})|x_t,c)}{p_\text{ref}(M(x_{t-1})|x_t,c)}\Big]\\ 
=&\underset{p_\theta}{\text{min}} E_{c,p_\theta(x_{0:T}|c)}\Big[ -\Sigma_{t=1}^{T} R_m(M(x_{t-1}),c)/\beta + \Sigma_{t=1}^{T}\text{log}\frac{p_\theta(M(x_{t-1})|x_t,c)}{p_\text{ref}(M(x_{t-1})|x_t,c)}\Big],\\ 
\end{aligned}
\end{equation}
where $r_m(x_0,c,M)$ is decomposed as $r_m(x_0,c,M)=E_{p_\theta(x_{1:T}|x_0,c)}[\Sigma_{t=0}^{T-1}R_m(M(x_{t}),c)]$, and $R_m(M(x_{t}),c)$ is the step-wise reward function.

Then, we formulate the sampling process of diffusion models under the multi-step RL framework as in \cite{rafailov2023direct}. We first define the initial state distribution $\rho_0$, state transition dynamics $P_s$, policy function $\pi$, action space $\mathcal{A}$, and the state space $\mathcal{S}$. Specifically, $P_s(s_{t+1}|s_t,a_t)$ takes the current state $s_t \in \mathcal{S}$ and action $a_t \in \mathcal{A}$ as input, returning the distribution of the next state $s_{t+1}$. The policy $\pi(a_t|s_t)$ determines the action for the current state. Different from previous literature, we propose the reward function $\hat{r}(s_t,\hat{a}_t)$, which receives $s_t$ and the action $\hat{a}_t$ from a subspace of $\mathcal{A}$. Then, we have:
% the tuple $(\rho_0, P_s, \mathcal{R},\mathcal{A}, \mathcal{S})$ for MDP below, which represents 
\begin{equation}
\begin{aligned}\label{eq:rl_defintion}
s_t \triangleq (x_t&,t,c) \\
\rho_0 \triangleq (\mathcal{N}(\mathbf{0},\mathbf{I}&),\delta_T,p(c)) \\
a_t \triangleq x&_{t-1} \\
\hat{a}_t \triangleq M(&x_{t-1}) \\
P_s(s_{t+1}|s_t,a_t) \triangleq &(\delta_{x_{t-1}},\delta_{t-1},\delta_c)\\
\hat{r}(s_t,\hat{a}_t)\triangleq R_m&(M(x_{t-1}),c), \\
\end{aligned}
\end{equation}
where $\delta$ is the Dirac delta function. Following \cite{levine2018reinforcement}, we can derive the optimal state-action value function $Q_m^*$, the optimal state value function $V_m^*$, and the optimal distribution $p_\theta^*$ from \cref{eq:mdpo_objective_2}:
\begin{equation}
\begin{aligned}\label{eq:first_derive_reward_objective_1}
Q^*_m((x_t,t,c),M(x_{t-1})) &= R_m(M(x_{t-1}),c) +E_{p^*_\theta(x_{t-1}|x_t,c,M(x_{t-1}))}[V_m^*(x_{t-1},t-1,c,M)],\\
V_m^*(x_{t},t,c,M)&=\beta\text{log}\int \text{exp}(Q^*((x_t,t,c),M(x_{t-1}))/\beta)d(M(x_{t-1})),\\
p_\theta^*(M(x_{t-1})|x_t,c)&={p_\text{ref}(M(x_{t-1})|x_t,c)\text{exp}\big((Q^*_m((x_t,t,c),M(x_{t-1}))-V_m^*(x_t,t,c,M))/\beta\big)}.
\end{aligned}
\end{equation}
Therefore, we have:
\begin{equation}
\begin{aligned}\label{eq:first_derive_reward_objective_2}
Q^*_m((x_t,t,c),M(x_{t-1}))&= \beta\text{log}\frac{p_\theta^*(M(x_{t-1})|x_t,c)}{p_\text{ref}(M(x_{t-1})|x_t,c)}+V_m^*(x_{t},t,c,M),\\
R_m(M(x_{t-1}),c)&=Q^*_m((x_t,t,c),M(x_{t-1}))-E[V_m^*(x_{t-1},t-1,c,M)] \\
&=\beta\text{log}\frac{p_\theta^*(M(x_{t-1})|x_t,c)}{p_\text{ref}(M(x_{t-1})|x_t,c)}+V_m^*(x_t,t,c,M)-E[V_m^*(x_{t-1},t-1,c,M)]. \\
\end{aligned}
\end{equation}
The total regional reward function is:
\begin{equation}
\begin{aligned}\label{eq:first_derive_reward_objective_3}
r_m(x_0,c,M)&=E_{p^*_\theta(x_{1:T}|x_0,c)}[\Sigma_{t=0}^{T-1}R_m(M(x_{t}),c)]\\
&=E_{p^*_\theta(x_{1:T}|x_0,c)}[\Sigma_{t=1}^{T}R_m(M(x_{t-1}),c)]\\
&= E_{p^*_\theta(x_{1:T}|x_0,c)}\Big[\Sigma_{t=1}^{T}\beta\text{log}\frac{p_\theta^*(M(x_{t-1})|x_t,c)}{p_\text{ref}(M(x_{t-1})|x_t,c)} + V_m^*(x_t,t,c,M)-E[V_m^*(x_{t-1},t-1,c,M)]\Big]\\
&=\beta TE_{t,x_{t-1,t}\sim p^*_\theta(x_{t-1,t}|x_0,c)}\Big[ \text{log}\frac{p_\theta^*(M(x_{t-1})|x_{t},c)}{p_\text{ref}(M(x_{t-1})|x_t,c)} \Big]+ \\
&\qquad \qquad E_{p^*_\theta(x_{1:T}|x_0,c)}[\Sigma_{t=1}^TV_m^*(x_t,t,c,M)-E[V_m^*(x_{t-1},t-1,c,M)]]\\
&=\beta TE_{t,x_{t-1,t}\sim p^*_\theta(x_{t-1,t}|x_0,c)}\Big[ \text{log}\frac{p_\theta^*(M(x_{t-1})|x_{t},c)}{p_\text{ref}(M(x_{t-1})|x_t,c)} \Big]+ Z_m(c,M),\\
\end{aligned}
\end{equation}
which is exactly the form of Eq. (8).

\subsection{Alternative Approximate Derivation of Eq. (8).}\label{sec:appendix_alter_derivation_8}
We also provide an alternative approximate derivation of Eq. (8). Start from \cref{eq:mdpo_objective_2}, we have:
\begin{equation}
\begin{aligned}\label{eq:derive_reward_objective_1}
&\underset{p_\theta}{\text{min}} E_{c,p_\theta(x_{0:T}|c)}\Big[ -\Sigma_{t=1}^{T} R_m(M(x_{t-1}),c)/\beta + \Sigma_{t=1}^{T}\text{log}\frac{p_\theta(M(x_{t-1})|x_t,c)}{p_\text{ref}(M(x_{t-1})|x_t,c)}\Big]\\
= &\underset{p_\theta}{\text{min}} E_{c,p_\theta(x_{0:T}|c)}\Big[ -\Sigma_{t=1}^{T} R_m(M(x_{t-1}),c)/\beta + \text{log}\frac{\prod_{t=1}^{T}p_\theta(M(x_{t-1})|x_t,c)}{\prod_{t=1}^{T}p_\text{ref}(M(x_{t-1})|x_t,c)}\Big]\\ 
= &\underset{p_\theta}{\text{min}} E_{c,p_\theta(x_{0:T}|c)} \Big[\text{log}\frac{\prod_{t=1}^{T}p_\theta(M(x_{t-1})|x_t,c)}{\prod_{t=1}^{T}p_\text{ref}(M(x_{t-1})|x_t,c)\text{exp}(R_m(M(x_{t-1}),c)/\beta)}\Big]\\ 
= &\underset{p_\theta}{\text{min}} E_{c,p_\theta(x_{0:T}|c)}\Big[ \Sigma_{t=1}^T\text{log}\frac{p_\theta(M(x_{t-1})|x_t,c)}{p_\text{ref}(M(x_{t-1})|x_t,c)\text{exp}(R_m(M(x_{t-1}),c)/\beta)}\Big]\\ 
= &\underset{p_\theta}{\text{min}} \Sigma_{t=1}^T E_{c,p_\theta(x_{t}|c)}E_{p_\theta(x_{t-1}|x_t,c)} \Big[\text{log}\frac{p_\theta(M(x_{t-1})|x_t,c)}{p_\text{ref}(M(x_{t-1})|x_t,c)\text{exp}(R_m(M(x_{t-1}),c)/\beta)}\Big]\\ 
= &\underset{p_\theta}{\text{min}} \Sigma_{t=1}^T E_{c,p_\theta(x_{t}|c)}E_{p_\theta(M(x_{t-1})|x_t,c)} \Big[\text{log}\frac{p_\theta(M(x_{t-1})|x_t,c)}{p_\text{ref}(M(x_{t-1})|x_t,c)\text{exp}(R_m(M(x_{t-1}),c)/\beta)/Z_m^t(c,M)}-\text{log}(Z_m^t(c,M))\Big]\\ 
= &\underset{p_\theta}{\text{min}} \Sigma_{t=1}^T E_{c,x_t}[\mathcal{D}_\text{KL} \big(p_\theta(M(x_{t-1})|x_t,c) || p_\text{ref}(M(x_{t-1})|x_t,c)\text{exp}(R_m(M(x_{t-1}),c)/\beta)/Z_m^t(c,M)\big)]-E[\text{log}(Z_m^t(c,M))],\\ 
\end{aligned}
\end{equation}
where $Z_m^t(c,M)=\Sigma_{x_{t-1}}p_\text{ref}(M(x_{t-1})|x_t,c)\text{exp}(R_m(M(x_{t-1}),c)/\beta)$ is the regional partition function for each step. Therefore, the optimal distribution $p_\theta^*(M(x_{t-1})|x_{t},c)$ for each step can be approximated as:
\begin{equation}
\begin{aligned}\label{eq:derive_reward_objective_2}
p_\theta^*(M(x_{t-1})|x_{t},c)&=p_\text{ref}(M(x_{t-1})|x_t,c)\text{exp}(R_m(M(x_{t-1}),c)/\beta)/Z^t_m(c,M). \\
\end{aligned}
\end{equation}
Then, we have:
\begin{equation}
\begin{aligned}\label{eq:derive_reward_objective_3}
R_m(M(x_{t-1}),c)&=\beta \text{log}\frac{p_\theta^*(M(x_{t-1})|x_{t},c)}{p_\text{ref}(M(x_{t-1})|x_t,c)}+\beta \text{log}Z^t_m(c,M). \\
\end{aligned}
\end{equation}
The total regional reward function is:
\begin{equation}
\begin{aligned}\label{eq:derive_reward_objective_4}
r_m(x_0,c,M)&=E_{p^*_\theta(x_{1:T}|x_0,c)}[\Sigma_{t=0}^{T-1}R_m(M(x_{t}),c)]\\
&=E_{p^*_\theta(x_{1:T}|x_0,c)}[\Sigma_{t=1}^{T}R_m(M(x_{t-1}),c)]\\
&=E_{p^*_\theta(x_{1:T}|x_0,c)}\Big[\beta \Sigma_{t=1}^T \text{log}\frac{p_\theta^*(M(x_{t-1})|x_{t},c)}{p_\text{ref}(M(x_{t-1})|x_t,c)}+\beta \Sigma_{t=1}^T\text{log}Z^t_m(c,M) \Big] \\
&=E_{p^*_\theta(x_{1:T}|x_0,c)}\Big[\beta \Sigma_{t=1}^T \text{log}\frac{p_\theta^*(M(x_{t-1})|x_{t},c)}{p_\text{ref}(M(x_{t-1})|x_t,c)}\Big] +\beta E_{p^*_\theta(x_{1:T}|x_0,c)}[\Sigma_{t=1}^T\text{log}Z^t_m(c,M)]\\
&=\beta TE_{t,x_{t-1,t}\sim p^*_\theta(x_{t-1,t}|x_0,c)}\Big[ \text{log}\frac{p_\theta^*(M(x_{t-1})|x_{t},c)}{p_\text{ref}(M(x_{t-1})|x_t,c)}\Big]+Z_m(c,M),\\
\end{aligned}
\end{equation}
which is exactly the form of Eq. (8).

\subsection{Derivation of Eq. (9)} The derivation of Eq. (9) is introduced below. We first substitute $r_m$ terms from Eq. (9) with Eq. (8): 
\begin{equation}
\begin{aligned}\label{eq:derive_mdpo_loss_1}
L_\text{R-DPO}&=-E_{c,x_0^w,x_0^l,M^w,M^l}\Big[\text{log}\sigma\big(r_m(x_0^w,c,M^w)-r_m(x_0^l,c,M^l) \big)\Big]\\
&=-E_{c,x_0^w,x_0^l,M^w,M^l}\Big[\text{log}\sigma\big(\beta TE_{t,x_{t}^w,x_{t}^l}E_{x_{t-1}^w,x_{t-1}^l}[\text{log}\frac{p_\theta(M^w(x_{t-1}^w)|x_{t}^w,c)}{p_\text{ref}(M^w(x_{t-1}^w)|x_{t}^w,c)} - \\
&\text{log}\frac{p_\theta(M^l(x_{t-1}^l)|x_{t}^l,c)}{p_\text{ref}(M^l(x_{t-1}^l)|x_{t}^l,c)}]+\Delta Z_m(c,M^w,M^l)\big)\Big], \\
\end{aligned}
\end{equation}
where $\Delta Z_m(c,M^w,M^l)$ takes the form $Z_m(c,M^w)-Z_m(c,M^l)$. By Jensen’s inequality, we have:
\begin{equation}
\begin{aligned}\label{eq:derive_mdpo_loss_2}
L_\text{R-DPO}&\leq-E_{c,t,x_0^w,x_0^l,M^w,M^l}E_{x_t^w,x_t^l}\Big[\text{log}\sigma\big(\beta TE_{x_{t-1}^w \sim q(x_{t-1}|x^w_{0,t}),x_{t-1}^l\sim q(x_{t-1}|x^l_{0,t})}[\text{log}\frac{p_\theta(M^w(x_{t-1}^w)|x_{t}^w,c)}{p_\text{ref}(M^w(x_{t-1}^w)|x_{t}^w,c)} - \\
&\text{log}\frac{p_\theta(M^l(x_{t-1}^l)|x_{t}^l,c)}{p_\text{ref}(M^l(x_{t-1}^l)|x_{t}^l,c)}]+\Delta Z_m(c,M^w,M^l)\big)\Big] \\
&=-E_{c,t,x_0^w,x_0^l,M^w,M^l}E_{x_t^w,x_t^l}\Big[\text{log}\sigma\big(\beta TE_{M^w(x_{t-1}^w) \sim q(M^w(x_{t-1})|x^w_{0,t}),M^l(x_{t-1}^l)\sim q(M^l(x_{t-1})|x^l_{0,t})} \\
&[\text{log}\frac{p_\theta(M^w(x_{t-1}^w)|x_{t}^w,c)}{p_\text{ref}(M^w(x_{t-1}^w)|x_{t}^w,c)} - \text{log}\frac{p_\theta(M^l(x_{t-1}^l)|x_{t}^l,c)}{p_\text{ref}(M^l(x_{t-1}^l)|x_{t}^l,c)}]+\Delta Z_m(c,M^w,M^l)\big)\Big], \\
\end{aligned}
\end{equation}
where we approximate $p_\theta$ with forward process $q$ as in \cite{wallace2024diffusion}. Then, we have:
\begin{equation}
\begin{aligned}\label{eq:derive_mdpo_loss_3}
L_\text{R-DPO}&\leq-E_{c,t,x_0^w,x_0^l,M^w,M^l}E_{x_t^w,x_t^l}\Big[\text{log}\sigma\big(\beta TE_{M^w(x_{t-1}^w),M^l(x_{t-1}^l)}[\text{log}p_\theta(M^w(x_{t-1}^w)|x_{t}^w,c)-\text{log}q(M^w(x_{t-1}^w)|x_{0,t}^w)- \\
&(\text{log}p_\text{ref}(M^w(x_{t-1}^w)|x_{t}^w,c)-\text{log}q(M^w(x_{t-1}^w)|x_{0,t}^w)) - (\text{log}p_\theta(M^l(x_{t-1}^l)|x_{t}^l,c)-\text{log}q(M^l(x_{t-1}^l)|x_{0,t}^l)- \\
&(\text{log}p_\text{ref}(M^l(x_{t-1}^l)|x_{t}^l,c)-\text{log}q(M^l(x_{t-1}^l)|x_{0,t}^l)))]+\Delta Z_m(c,M^w,M^l)\big)\Big] \\
&=-E_{c,t,x_0^w,x_0^l,M^w,M^l}E_{x_t^w,x_t^l}\Big[\text{log}\sigma \Big(-\beta T\big(\mathcal{D}_\text{KL}\big(q(M^w(x^w_{t-1})|x^w_{0,t}) || p_\theta(M^w(x^w_{t-1})|x^w_{t},c)\big)-\\
&\mathcal{D}_\text{KL}\big(q(M^w(x^w_{t-1})|x^w_{0,t}) || p_\text{ref}(M^w(x^w_{t-1})|x^w_{t},c)\big) -\big(\mathcal{D}_\text{KL}\big(q(M^l(x^l_{t-1})|x^l_{0,t}) || p_\theta(M^l(x^l_{t-1})|x^l_{t},c)\big)- \\
&\mathcal{D}_\text{KL}\big(q(M^l(x^l_{t-1})|x^l_{0,t}) || p_\text{ref}(M^l(x^l_{t-1})|x^l_{t},c)\big)\big)\big)+ \Delta Z_m(c,M^w,M^l)\Big)\Big]. \\
\end{aligned}
\end{equation}
From previous literature~\cite{ho2020denoising},  we have:
\begin{equation}
\begin{aligned}\label{eq:derive_mdpo_loss_4}
E_{c,t,x_0}E_{x_t}[\text{D}_{\text{KL}}(q(x_{t-1}|x_{0,t}) || p_\theta(x_{t-1}|x_{t},c))]= E_{c,t,x_0,\epsilon}[\omega_t||\epsilon-\epsilon_\theta(x_t,t,c)||^2],\\
\end{aligned}
\end{equation}
where $p_\theta(x_{t-1}|x_t,c)$ takes the form $\mathcal{N}(x_{t-1}|\mu_\theta,\sigma_t^2\mathbf{I}$), and $q(x_{t-1}|x_{0,t}):=\mathcal{N}(x_{t-1}|\mu_t,\sigma_t^2\mathbf{I}$). Since $M$ is a linear transformation, $p_\theta(M(x_{t-1})|x_{t},c)$ and $q(M(x_{t-1})|x_{0,t})$ become to $\mathcal{N}(x_{t-1}|M(\mu_\theta),\sigma_t^2\mathbf{I})$ and $\mathcal{N}(x_{t-1}|M({\mu}_t),\sigma_t^2\mathbf{I})$, respectively. Therefore, we can extend the \cref{eq:derive_mdpo_loss_4} into following regional variant:
\begin{equation}
\begin{aligned}\label{eq:derive_mdpo_loss_5}
E_{c,t,x_0,M}E_{x_t}[\text{D}_{\text{KL}}(q(M(x_{t-1})|x_{0,t}) || p_\theta(M(x_{t-1})|x_{t},c))]= E_{c,t,x_0,M,\epsilon}[\omega_t||M(\epsilon-\epsilon_\theta(x_t,t,c))||^2],\\
\end{aligned}
\end{equation}
Since $||v-v_\theta(x_t,t,c)||^2 \propto ||\epsilon-\epsilon_\theta(x_t,t,c)||^2$ ~\cite{liu2025improving}, we can get following objective by substituting the \cref{eq:derive_mdpo_loss_5} into each KL-divergence term of \cref{eq:derive_mdpo_loss_3}:
\begin{equation}
\begin{aligned}\label{eq:derive_mdpo_loss_6}
L_\text{R-DPO}&\leq-E_{c,t,x_0^w,x_0^l,M^w,M^l,\epsilon^w,\epsilon^l}\Big[\text{log}\sigma\Big(-\beta T \omega_t\big(||M^w(v^w-v_\theta(x^w_t,t,c))||^2_2-||M^w(v^w-v_\text{ref}(x^w_t,t,c))||^2_2-\\
&(||M^l(v^l-v_\theta(x^l_t,t,c))||^2_2-||M^l(v^l-v_\text{ref}(x^l_t,t,c))||^2_2)\big)+\Delta Z_m(c,M^w,M^l) \Big)\Big], \\
\end{aligned}
\end{equation}
which is exactly the form of Eq. (9). According to the formulation of $Z_m(c,M)$ from \cref{eq:first_derive_reward_objective_3} and \cref{eq:derive_reward_objective_4}, $\Delta Z_m(c,M^w,M^l)$ equals $0$ when $M^w=M^l$.

\subsection{Comparison of \dpomethodname with Previous Methods}
Recently, several studies~\cite{wu2025densedpo,gu2025mask} have introduced their DPO variants for video generation models and Large Language Models, to address the inefficiency of applying overall preferences~\cite{rafailov2023direct} in their tasks. For example, DenseDPO~\cite{wu2025densedpo} assigns preference labels to each frame for two video samples, while Mask-DPO~\cite{gu2025mask} leverages sentence-level preference annotations to enable the model to only learn from the correct facts in winning answers and the incorrect contents in losing answers. In contrast, our \dpomethodname is proposed for image generation and constructs region-level preference pairs within the spatial dimension. Moreover, while providing a comprehensive derivation, we extend the conventional single preference pair to a group-wise setting, thereby enhancing sample diversity and data utilization efficiency.

For image generation, the most related work to ours is PatchDPO~\cite{huang2025patchdpo}. Similarly, it introduces a patch-level DPO objective for customization tasks to improve subject consistency. However, our \dpomethodname differs from the method in the following aspects. \textbf{1).} Essentially, PatchDPO employs a weighted diffusion loss, which can be regarded as a softer variant of our Mask-SFT discussed in the ablation studies. Specifically, it assigns high/low weights to superior/inferior patches within an image, with weights normalized to $[0,1]$. However, the objective inherently lacks explicit penalty signals from losing samples, leading to suboptimal model performance, as discussed in our ablation studies. In contrast, our \dpomethodname aligns the model outputs with winning glyph regions, while distancing them from losing ones, thereby learning localized glyph correctness. \textbf{2).} Similar to previous works~\cite{wu2025densedpo,rafailov2023direct,gu2025mask}, PatchDPO objective only considers a single preference pair for each batch, which limits the sample diversity. Our \dpomethodname objective, on the other hand, generates a group of images per condition, allowing the model to learn from the superior regions across different samples and thereby improving overall model performance.

\section{More Details of \infermethodname} 
First, we prove that Eq.~(12) derives the optimal distribution with adjustable regularization weight. As shown in \cref{eq:first_derive_reward_objective_1} and \cref{eq:derive_reward_objective_2}, since $p_\theta(x_t|c) \propto p_\text{ref}(x_t|c)\text{exp}(r(x_t,c)/\beta)$, we can reformulate Eq. (12) as :
\begin{equation}
\begin{aligned}\label{eq:derive_dpo_cfg}
\hat{s}_{\text{ref},\theta}(x_t,\omega,c)&=\nabla_{x_t}\text{log}\big(p_\text{ref}(x_t|c)^{(1-\omega)}p_\theta(x_t|c)^\omega\big) \\
&=\nabla_{x_t}\text{log}\big(p_\text{ref}(x_t|c)^{(1-\omega)}(p_\text{ref}(x_t|c)\text{exp}(r(x_t,c)/\beta))^\omega \big) \\
&=\nabla_{x_t}\text{log}\Big(p_\text{ref}(x_t|c)\text{exp}(\frac{r(x_t,c)}{\beta/\omega}) \Big). \\
\end{aligned}
\end{equation}
The $p_\text{ref}(x_t|c)\text{exp}(\frac{r(x_t,c)}{\beta/\omega})$ in the last row corresponds to the optimal distribution to be sampled. $\beta/\omega$ is the adjustable regularization weight, controlling the glyph accuracy during inference. Then, we give the full derivation of Eq.~(13):
\begin{equation}
\begin{aligned}\label{eq:derive_dpo_cfg_2}
v^*(x_t,t,c)&= a_tx+b_t \big(\nabla_{x_t}\text{log}\big(p_\text{ref}(x_t|c)^{(1-\omega)}p_\theta(x_t|c)^\omega\big)\big)\\
&= a_tx+b_t \big((1-\omega)\nabla_{x_t}\text{log}p_\text{ref}(x_t|c)+\omega\nabla_{x_t}\text{log}p_\theta(x_t|c)\big)\\
&= a_tx+b_t \big((1-\omega)\frac{v_\text{ref}(x_t,t,c)-a_tx}{b_t}+\omega\frac{v_\theta(x_t,t,c)-a_tx}{b_t}\big)\\
&= a_tx+(1-\omega)v_\text{ref}(x_t,t,c)+\omega v_\theta(x_t,t,c)-a_tx\\
&= (1-\omega)v_\text{ref}(x_t,t,c)+\omega v_\theta(x_t,t,c).\\
\end{aligned}
\end{equation}
\textbf{Algorithm} \ref{alg:infer_method} demonstrates the pipeline of our \infermethodname. 
\begin{algorithm}[h]
\small
\caption{Algorithm of \infermethodname.}
\LinesNumbered
\label{alg:infer_method}
\textbf{Input:} Initial noisy image $x_T$ sampled from $\mathcal{N}(\mathbf{0},\mathbf{I})$; condition $c$; Classifier-Free Guidance scale $\omega$; model $v_{\text{ref}}$ from Stage 1; model $v_{\theta}$ from Stage 2; sampling step $T$;\\
\textbf{Output:} The generated image $x$;\\
Get the overall text region mask $\hat{\mathbf{P}}$ from $c$ ;\\
\For{$t~\mathrm{in}~\{T,T-1 \ldots, 1\}$}
{
$v_{t,\text{ref}}=v_{\text{ref}}(x_{t},t, c)$; \\
$v_{t,\theta}=v_{\theta}(x_{t},t, c)$; \\
$v_t^{*}=(1-\omega)v_{t,\text{ref}} +\omega v_{t,\theta}$; \\
$\hat{v}_t^{*}=\hat{\mathbf{P}}v_{t}^{*}+(\mathbf{I}-\hat{\mathbf{P}}) v_{t,\theta}$; \\
$x_{t-1}=x_{t}+\hat{v}_t^* dt$; \\
}
$x=x_0$;\\
\textbf{Return:} The generated image $x$.  
\end{algorithm}

\begin{figure*}
    \centering
    \includegraphics[width=1\textwidth]{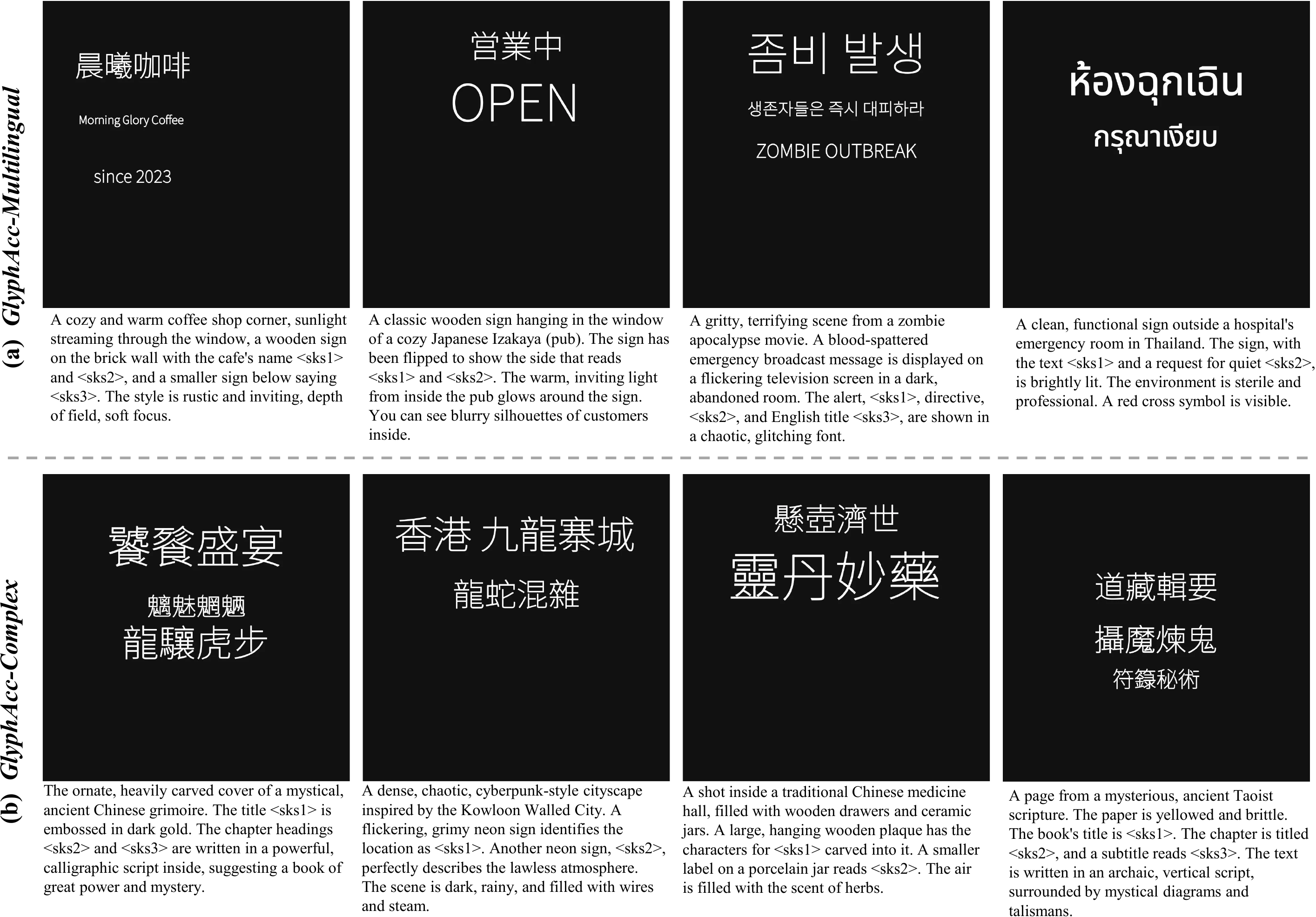}
    \caption{Samples from the constructed benchmarks \textbf{(a)} \multilingualbenchmarkname and \textbf{(b)} \complexbenchmarkname.}
    \label{fig:appendix-benchmark}
\end{figure*}

\begin{figure*}[h]
    \centering
    \includegraphics[width=1\textwidth]{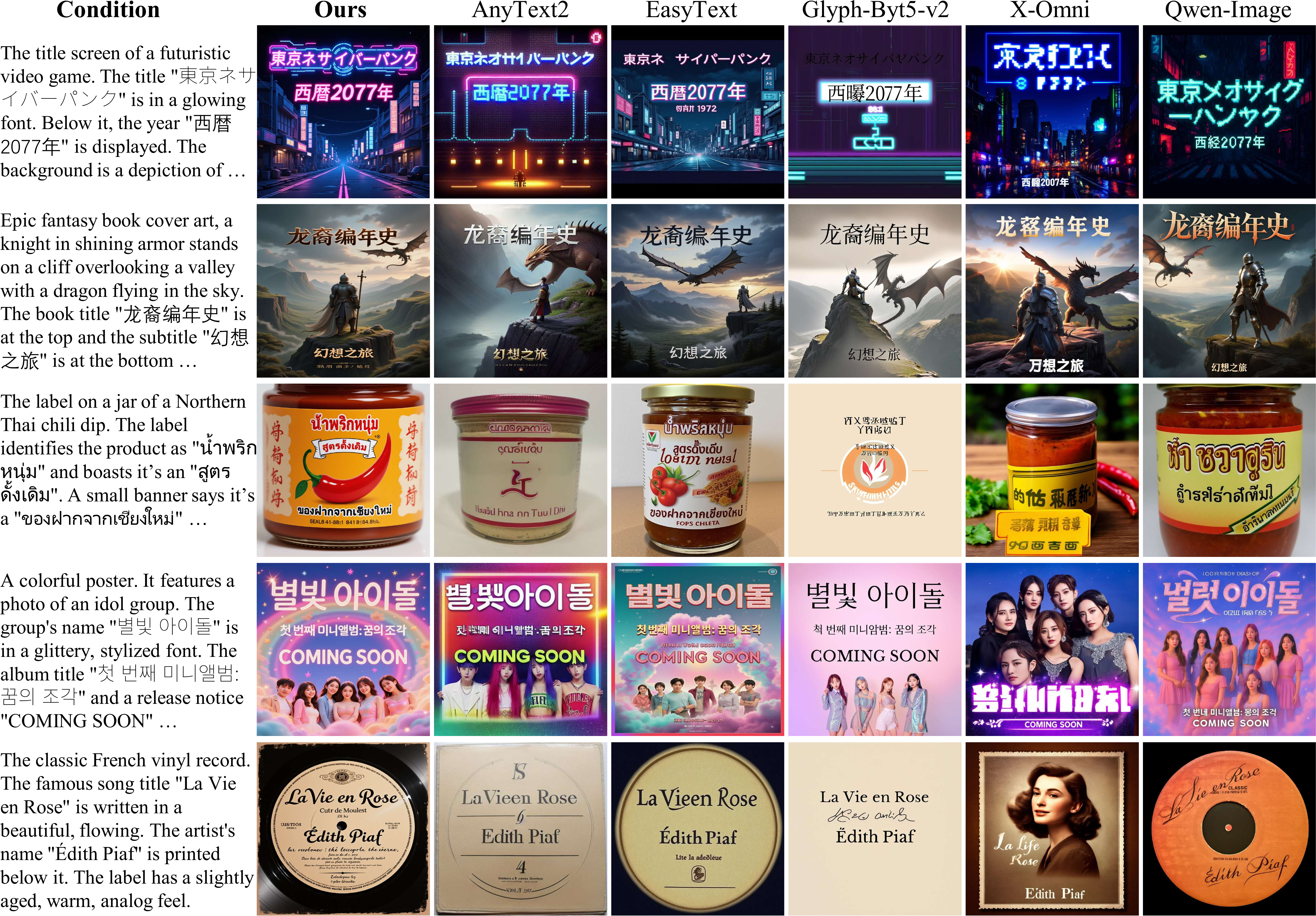}
    \caption{Comparison results on \multilingualbenchmarkname.}
    \label{fig:appendix_multilingual}
\end{figure*}

\paragraph{Comparison of \infermethodname With Previous Methods.} The early work Flow-NRG~\cite{liu2025improving} introduces reward guidance in inference time, enhancing model performance by sampling from the optimal distribution. Inspired by the classifier-guidance method~\cite{dhariwal2021diffusion}, Flow-NRG trains an additional reward network to assess noisy images. During inference, it leverages the gradient of the reward network to modulate the velocity field, guiding the sampling process. In contrast, our method builds upon Classifier-Free Guidance (CFG)~\cite{ho2022classifier}, which combines the predicted velocity fields from models at different stages, enabling sampling from a controllable optimal distribution without training an auxiliary network.

Inspired by CFG, some concurrent works~\cite{frans2025diffusion,cheng2025diffusion} have also introduced their reward guidance approaches. While they arrive at similar conclusions, our \infermethodname is derived by a different approach. Furthermore, we extend the conventional image-level guidance to the region-level variant, which better preserves the quality of the background content.

\section{More Details of Benchmarks}
We construct two benchmarks, \multilingualbenchmarkname and \complexbenchmarkname, to evaluate the model performance in rendering multilingual texts and characters with complex glyphs, respectively. \multilingualbenchmarkname consists of 370 test cases spanning seven languages: English, Chinese, Japanese, Korean, French, Vietnamese, and Thai. In contrast, \complexbenchmarkname contains 97 test cases, focusing on complex Chinese characters. For both benchmarks, we instruct Gemini~\cite{team2023gemini} to provide the image description and the coordinates of 1 to 4 bounding boxes for placing texts. Furthermore, the texts to be rendered are also obtained via Gemini. Representative examples are shown in \cref{fig:appendix-benchmark}. The prompt for querying Gemini is:  

\begin{quote}
You are an image layout expert, specializing in designing which areas of a 1024$\times$1024 image should contain text. You will output rectangular text bounding boxes to indicate the positions of the text in the image, in the format:  
\[(x_1,y_1),(x_2,y_2),(x_3,y_3),(x_4,y_4)\] (top-left, top-right, bottom-right, bottom-left).  
For each image, output 1--4 text bounding boxes. The content inside each bounding box should be text in English, Chinese, Japanese, Korean, French, Vietnamese, or Thai.  
Finally, provide an overall prompt for the image, using placeholders such as \texttt{<sks1>}, \texttt{<sks2>}, \ldots, \texttt{<sksn>} to denote the text that will be rendered.  
Output the result in a JSON file.  
\end{quote}

\begin{figure*}[b]
    \centering
    \includegraphics[width=1\textwidth]{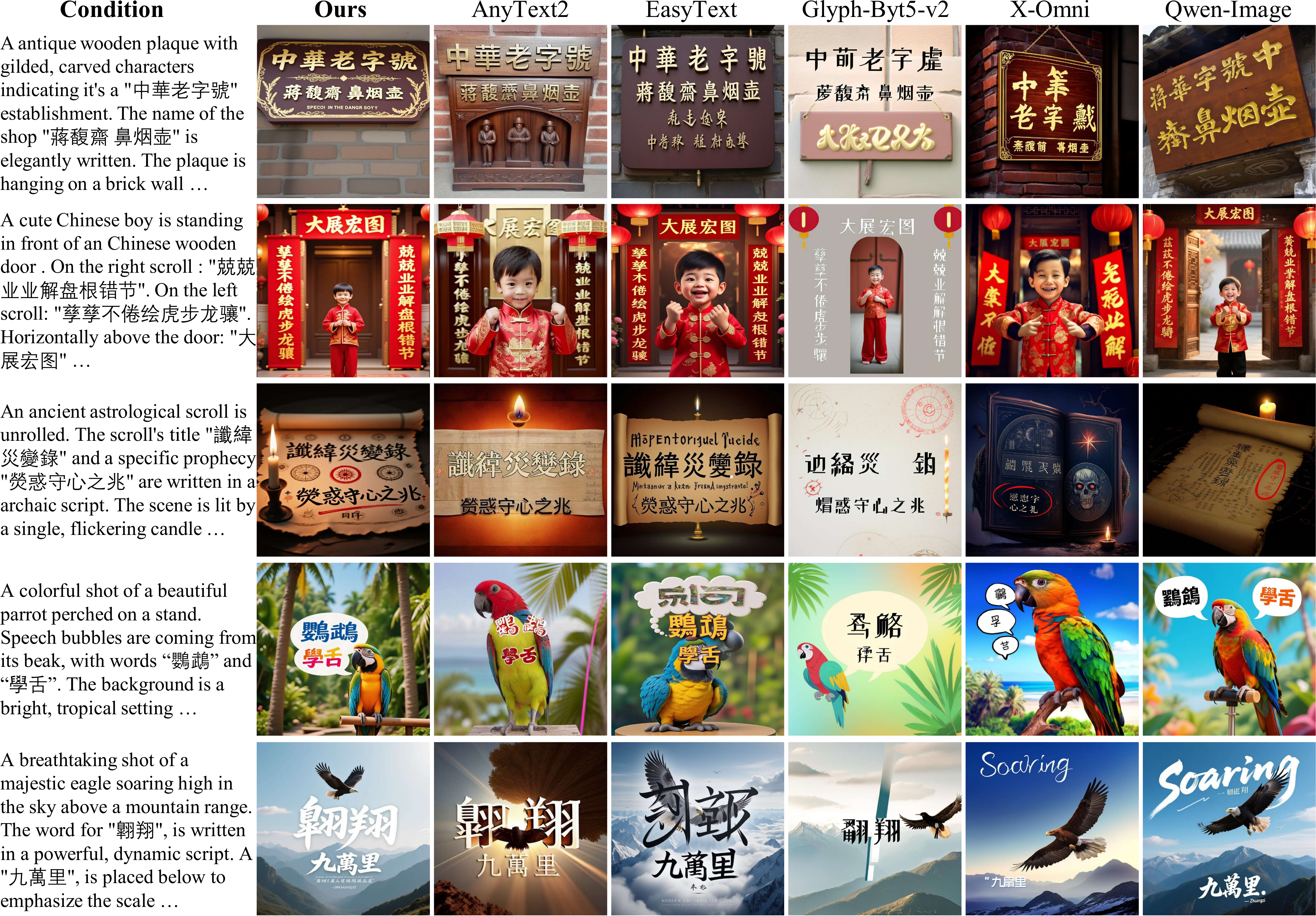}
    \caption{Comparison results on \complexbenchmarkname.}
    \label{fig:appendix_complex}
\end{figure*}
\section{More Details of Evaluation Metrics}
We leverage Qwen2.5-VL~\cite{bai2025qwen2} as VLM to evaluate the following metrics. 

\noindent\textbf{1. Text accuracy.}  
To calculate text accuracy metrics, such as Normalized Edit Distance (\textbf{NED}) and sentence accuracy (\textbf{Sen.Acc}), we first instruct VLM to recognize texts through the following query prompt:  
\begin{verbatim}
You are an expert in text recognition. Please recognize the text in the image and 
output it line by line.
\end{verbatim}

\noindent\textbf{2. Image aesthetic and text-image alignment.}  
We also employ VLM~\cite{bai2025qwen2} to evaluate image aesthetic (\textbf{Aes}) and text-image alignment (\textbf{Text.Align}), using the following query prompt:  
\begin{verbatim}
Please evaluate the provided text-image pair according to the following two criteria. 
The input consists of a prompt and a generated image.  
1. Image Aesthetic: Assess the visual quality of the image, including factors such as 
   color harmony, contrast, and overall appeal.  
2. Text-Image Alignment: Evaluate how well the generated image aligns with the given 
prompt. For each criterion, provide:  
- A score from 1 to 100 (where 1 = poor, 100 = excellent).  
- A brief explanation justifying the score.  
Return your evaluation in the following JSON-like format:  
{
  "Image Aesthetic": {
    "score": <score>,
    "comment": "<explanation>"
  },
  "Text-Image Alignment": {
    "score": <score>,
    "comment": "<explanation>"
  }
}
\end{verbatim}

\section{More Details of the User Study}  
We employ 20 volunteers to conduct the user study with the generated images from \multilingualbenchmarkname and \complexbenchmarkname benchmarks. They are asked to assess each image in terms of image aesthetic, text-image alignment, and glyph accuracy. For glyph accuracy, the participants need to compare the rendered characters in the glyph image with the generated ones. All of the above scores are within the range of 1 to 10. For each score, we average the results.

\begin{figure*}[t]
    \centering
    \includegraphics[width=1\textwidth]{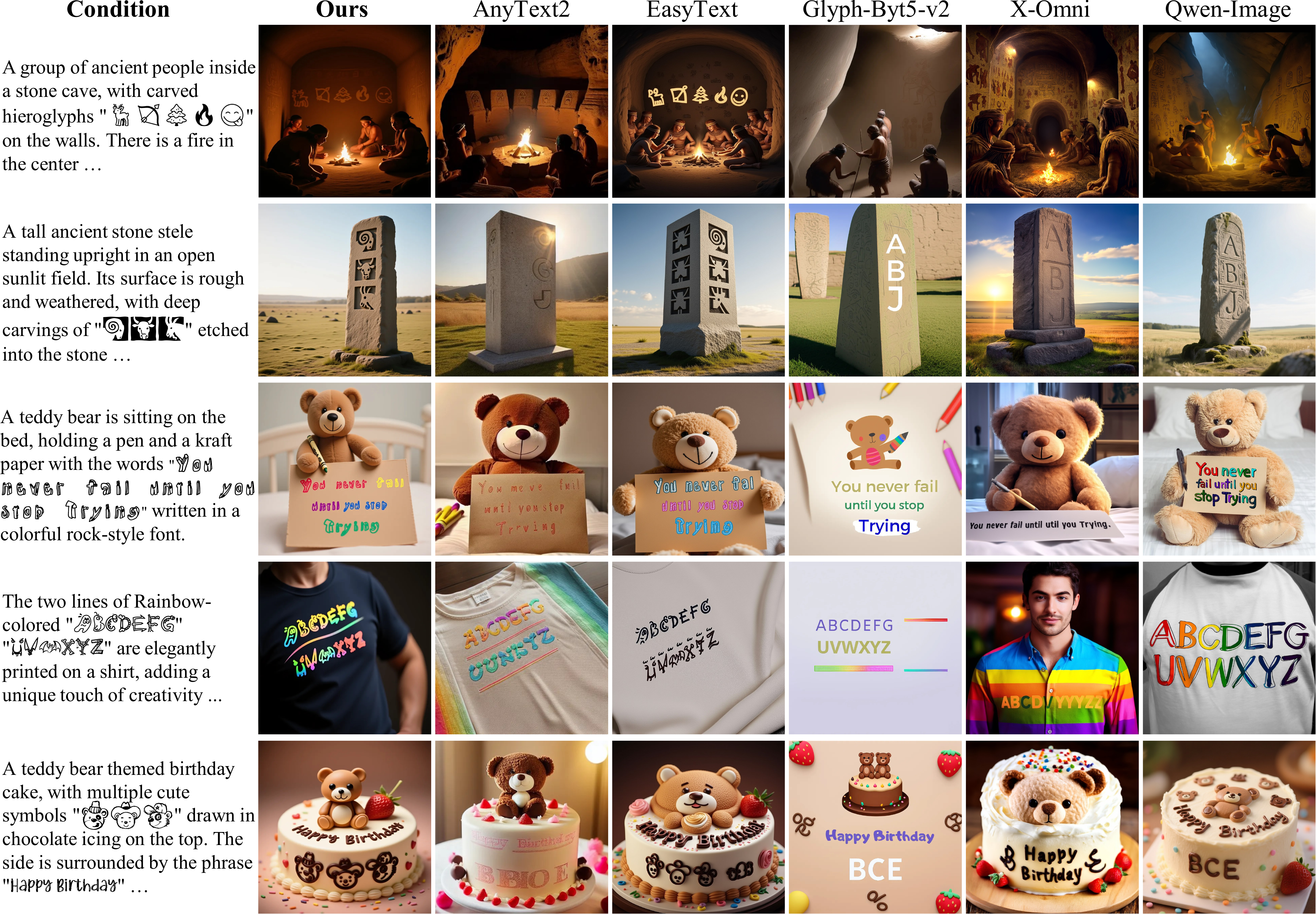}
    \caption{Comparison results under out-of-domain conditions.}
    \label{fig:appendix_symbol}
\end{figure*}

\begin{figure*}[t]
    \centering
    \includegraphics[width=1\textwidth]{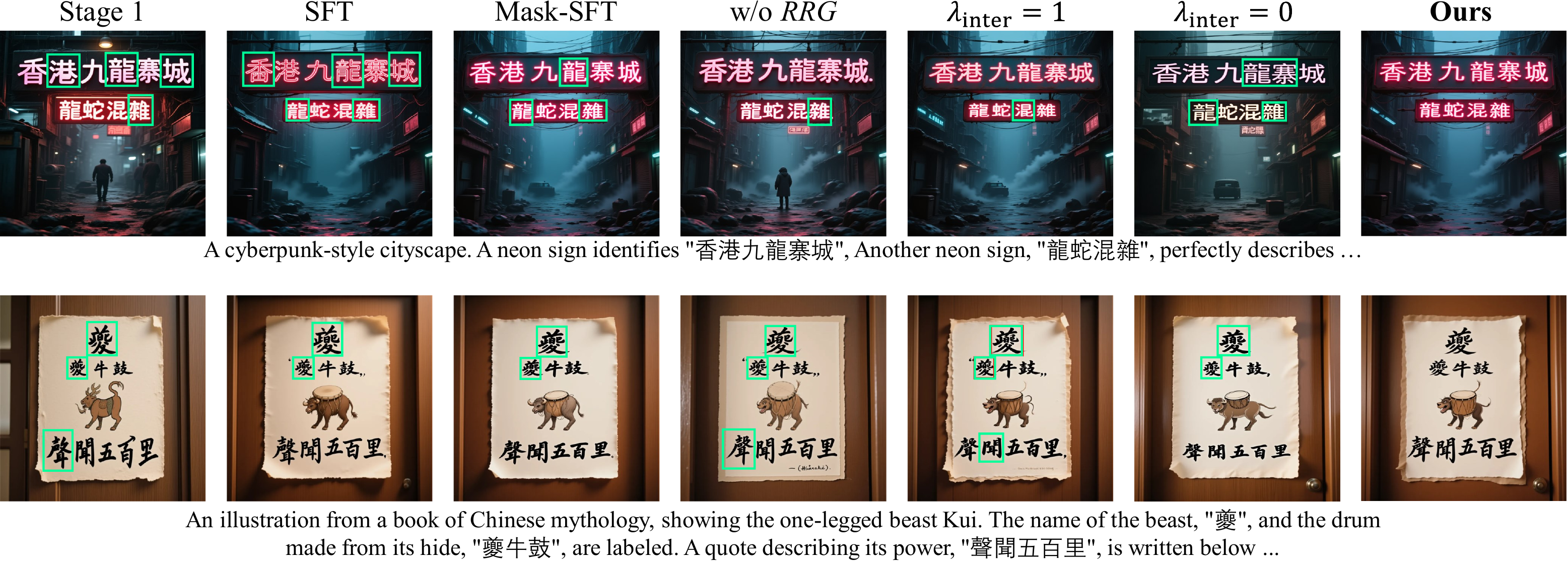}
    \caption{More results of ablation studies. The erroneous regions of each image are highlighted with green boxes.}
    \label{fig:appendix_ablation}
\end{figure*}

\section{More Comparison Results}
We provide more comparison results in this section. \cref{fig:appendix_multilingual} and \cref{fig:appendix_complex} represent the results from \multilingualbenchmarkname and \complexbenchmarkname, respectively.  As shown in \cref{fig:appendix_multilingual}, most of \textit{prompt-guided} methods fail to generate accurate glyphs for some infrequent languages, as shown in the 3rd row of the figure. For the examples from \cref{fig:appendix_complex}, existing methods exhibit poor performance in rendering complex glyphs. In contrast, our \methodname outperforms in glyph accuracy across all these cases, achieving a good balance between the stylization and precision.

In addition, we also illustrate the model performance on out-of-domain inputs, where we construct glyph conditions with emojis and characters with stylized fonts. As shown in \cref{fig:appendix_symbol}, only our method preserves fine-grained structural details of the condition.

\cref{fig:appendix_ablation} presents more results of ablation studies, demonstrating the effects of our key designs.

We further evaluate the performance of our \methodname against other comparison methods on the OneIG benchmark~\cite{chang2025oneig}. We first leverage Gemini~\cite{team2023gemini} to generate the corresponding layouts using the query prompt below.

\begin{quote}\label{quote:oneig}
You are an image layout expert, specializing in designing which areas of a 1024$\times$1024 image should contain text. You will output rectangular text bounding boxes to indicate the positions of the text in the image, in the format:  
\[(x_1,y_1),(x_2,y_2),(x_3,y_3),(x_4,y_4)\] (top-left, top-right, bottom-right, bottom-left).  
For each image, generate bounding boxes corresponding to the number of sentences that need to be rendered from the prompt. The content within each bounding box should be taken directly from the quoted prompt. You may split a sentence into smaller segments if it is too long, but each split segment must also generate its own independent bounding box. 
Finally, provide an overall prompt for the image, using placeholders such as \texttt{<sks1>}, \texttt{<sks2>}, \ldots, \texttt{<sksn>} to denote the text that will be rendered. 
Output the result in a JSON file.  
\end{quote}

As shown in \cref{tab:benchmark_comparison_supp}, our method achieves the best performance on both English and Chinese scenarios.

\begin{table*}[h]
    \centering

    \caption{The quantitative results on the OneIG~\cite{chang2025oneig} benchmark.}
    \vspace{-2mm}
    \label{tab:benchmark_comparison_supp}
    \scriptsize
    \setlength{\tabcolsep}{2pt}
    \renewcommand{\arraystretch}{1.3}
    \resizebox{\textwidth}{!}{
        \begin{tabular}{c|cc|cc|cc|cc|cc|cc}
            \toprule
            \textbf{Language} & \multicolumn{2}{c|}{\textbf{\methodname (Ours)}} & \multicolumn{2}{c|}{AnyText2~\cite{tuo2024anytext2}} & \multicolumn{2}{c|}{EasyText~\cite{lu2025easytext}} & \multicolumn{2}{c|}{Glyph-Byt5-v2~\cite{liu2024glyphv2}} & \multicolumn{2}{c|}{X-Omni~\cite{geng2025x}} & \multicolumn{2}{c}{Qwen-Image~\cite{wu2025qwen}} \\
            \hline
            & \textbf{NED} & \textbf{Sen.Acc} & \textbf{NED} & \textbf{Sen.Acc} & \textbf{NED} & \textbf{Sen.Acc} & \textbf{NED} & \textbf{Sen.Acc} & \textbf{NED} & \textbf{Sen.Acc} & \textbf{NED} & \textbf{Sen.Acc} \\
            \hline
            English & \textbf{0.9704} & \textbf{0.8853} & 0.6314 & 0.5301 & \underline{0.9571} & \underline{0.8741} & 0.8060 & 0.7650 & 0.8930 & 0.6353 & 0.9432 & 0.8327 \\
            Chinese & \textbf{0.9771} & \textbf{0.8932} & 0.7642 & 0.5089 & \underline{0.9589} & \underline{0.8808} & 0.9287 & 0.8274 & 0.7705 & 0.4199 & 0.9424 & 0.8630 \\
            \bottomrule
        \end{tabular}
    }
\end{table*}

}
\end{document}